\documentclass[letterpaper]{article}
\usepackage{ceavadpaper}
\nocopyright
\usepackage[hyphens]{url}
\usepackage{graphicx}
\usepackage{natbib}
\usepackage{caption}
\usepackage{algorithm}
\usepackage{algorithmic}
\usepackage{newfloat}
\usepackage{listings}
\DeclareCaptionStyle{ruled}{labelfont=normalfont,labelsep=colon,strut=off}
\floatstyle{ruled}
\newfloat{listing}{tb}{lst}{}
\floatname{listing}{Listing}
\usepackage{booktabs}
\usepackage[hidelinks]{hyperref}

\hypersetup{
  pdftitle={Beyond Hazard Resemblance: Contrastive Event Adjudication for Training-Free Video Anomaly Detection},
  pdfauthor={Wenti Yin, Xiang Wang, Huaxin Zhang, Hanqing Wang, Hongbo Shao, Changxin Gao, Nong Sang},
  pdfsubject={Video Anomaly Detection}
}

\title{Beyond Hazard Resemblance: Contrastive Event Adjudication for Training-Free Video Anomaly Detection}
\author{
  Wenti Yin\textsuperscript{\rm 1},
  Xiang Wang\textsuperscript{\rm 3},
  Huaxin Zhang\textsuperscript{\rm 1},
  Hanqing Wang\textsuperscript{\rm 2},
  Hongbo Shao\textsuperscript{\rm 1},\\
  Changxin Gao\textsuperscript{\rm 1},
  Nong Sang\textsuperscript{\rm 1}\thanks{Corresponding author.}
}
\affiliations{
  \textsuperscript{\rm 1}Key Laboratory of Image Processing and Intelligent Control,
  School of Artificial Intelligence and Automation,\\
  Huazhong University of Science and Technology\\
  \textsuperscript{\rm 2}The Hong Kong University of Science and Technology (Guangzhou)\\
  \textsuperscript{\rm 3}Alibaba Group\\
  \{yinwt, nsang\}@hust.edu.cn
}

\begin{document}

\maketitle

\begin{abstract}
Video anomaly detection (VAD) aims to identify and temporally localize abnormal events in videos. 
Supervised methods learn anomaly decision boundaries from target-domain annotations but require substantial 
in-domain data. Existing training-free methods leverage the rich semantic knowledge and reasoning capabilities of 
pretrained models to interpret visual content, yet these capabilities do not directly define an anomaly decision criterion: 
richer anomaly descriptions better capture hazard resemblance without resolving abnormality. 
To this end, we propose Contrastive Event Adjudication for training-free Video Anomaly Detection (CEAVAD), which shifts the unit of inference from 
isolated anomaly concepts to falsifiable event hypotheses and establishes an inference-time explanatory boundary through the 
interaction between competing explanations and video evidence. 
Specifically, CEAVAD first uses public-safety knowledge to construct hazard–benign event contrasts, 
pairing each hazard mechanism with a generic normal account and a mechanism-specific benign counterpart. 
It then determines whether the target interval better supports a hazard explanation or its benign competitor, 
yielding a revisable contrastive boundary proposal for the target. 
Finally, CEAVAD adjudicates between the competing explanations to determine whether the hazard hypothesis survives the video evidence, 
supporting both temporally localized anomaly detection and evidence-grounded explanations. Experiments on three widely used VAD benchmarks demonstrate 
that CEAVAD achieves state-of-the-art performance under the training-free paradigm. Code is available at \url{https://github.com/lessiYin/CEAVAD}.

\end{abstract}

\section{Introduction}
\label{sec:introduction}

Video anomaly detection (VAD) seeks to identify events that depart from expected behavior
and localize them along the video timeline.  Classical one-class and unsupervised methods
learn regularity from target-domain normal video, using reconstruction, prediction, or
memory-based discrepancy as anomaly evidence
\cite{hasan2016learning,liu2018future,park2020learning}.  Weakly supervised methods instead
learn a discriminative boundary from video-level normal and anomalous labels
\cite{sultani2018real,tian2021rtfm,wu2024vadclip}.  Although these paradigms differ in their
supervision, both encode the anomaly criterion in parameters fitted to data from the target
domain.  Deploying the detector in a new environment therefore entails collecting suitable
video, defining supervision, or optimizing the model again.

\begin{figure}[t]
    \centering
    \includegraphics[width=\columnwidth]{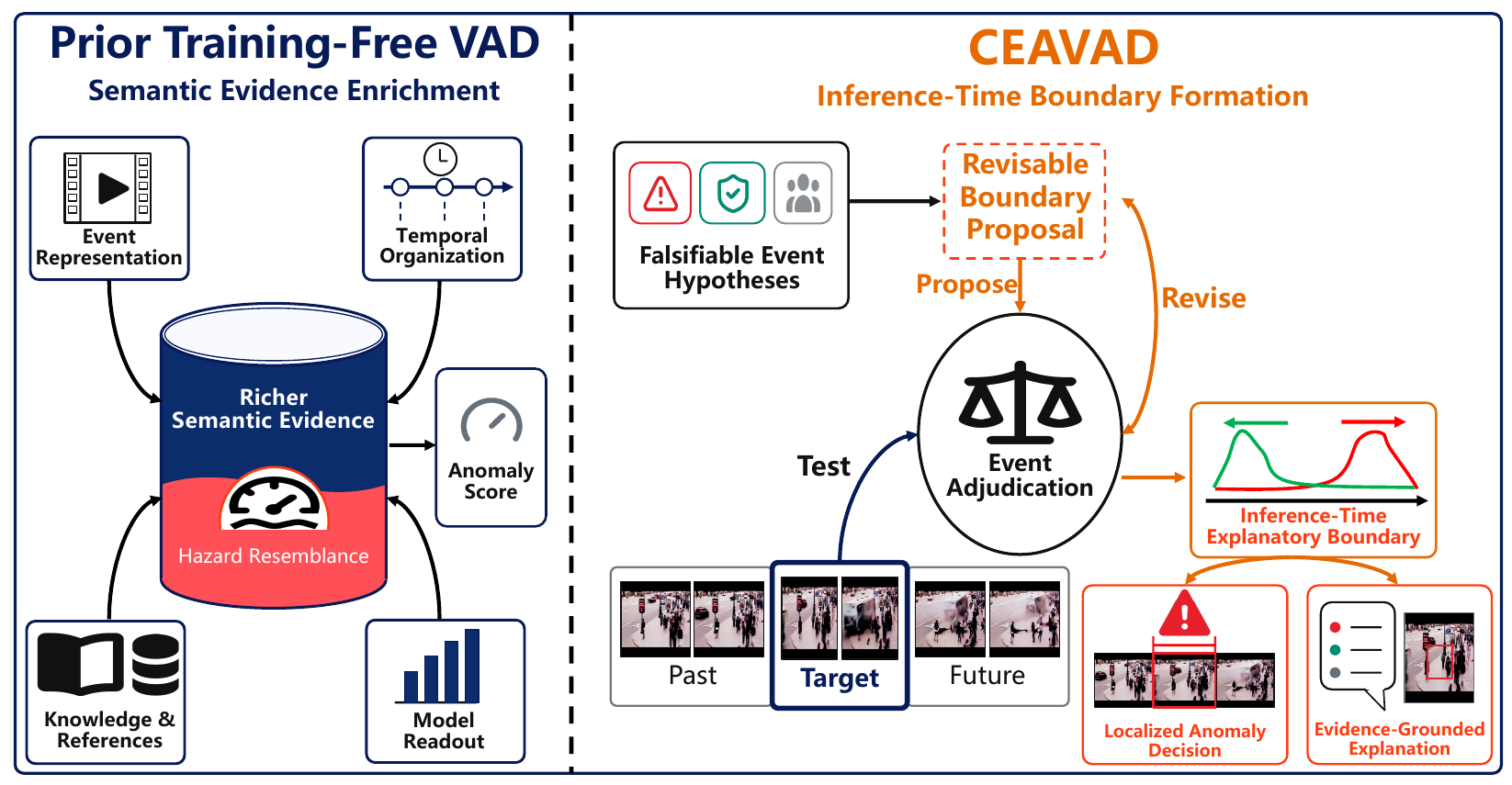}
    \caption{Schematic illustration of the motivation for CEAVAD.  Prior training-free VAD enriches
    semantic evidence, yet semantic evidence alone does not define the anomaly criterion.
    CEAVAD uses semantic contrast to form a revisable proposal and adjudicates falsifiable
    event hypotheses against temporally partitioned video evidence, establishing an
    inference-time explanatory boundary for both localized anomaly decisions and
    evidence-grounded explanations.}
    \label{fig:motivation}
\end{figure}

Pretrained foundation models offer a different route.  Vision--language pretraining exposes
broad visual semantics through natural language \cite{radford2021learning}, while large
language models (LLMs) and multimodal large language models (MLLMs) provide world knowledge
and event-level reasoning
without fitting a conventional VAD classifier.  Recent training-free methods pursue two
broad aims: enriching event evidence through multimodal descriptions, temporal organization,
and normal exemplars
\cite{zanella2024harnessing,dev2024mcanet,shao2025eventvad,li2025vadtree,sun2026rag4vad},
and strengthening anomaly judgment through induced normality rules, anomaly definitions,
and anomaly-sensitive representations
\cite{yang2024follow,pei2026drvad,cai2025hiprobe,cai2026headhunt}.

These advances improve semantic understanding, yet semantic knowledge alone does not define
an anomaly decision boundary.  A foundation model may recognize hazard resemblance, but VAD
must still determine whether the observed event is abnormal.  Hazard support therefore has
to be weighed against a generic normal account and a mechanism-specific benign counterpart
that shares the hazard's visual pattern.  The distinction is between evidence and criterion:
representations, references, and reasoning can enrich evidence, whereas the criterion
specifies how an interpretation is proposed, tested, and revised against temporally grounded
evidence.  Testing and revising candidate interpretations against such evidence establishes
an inference-time explanatory boundary for the abnormality judgment.
Figure~\ref{fig:motivation} illustrates this boundary-forming interaction.

Our key observation is that this interaction becomes testable by changing the unit of
reasoning from an isolated anomaly concept to a falsifiable event hypothesis.  A hazard
hypothesis should compete simultaneously with a generic normal account and a
mechanism-specific benign counterpart that can explain the same observation.  Their relative
semantic support forms a revisable proposal, while temporally organized video evidence can
confirm the hazard, reject it in favor of a benign explanation, or show that the hypothesized
event does not occur.  The resulting anomaly decision is grounded in which competing
explanation survives the visible evidence.
Based on this view, we introduce \emph{Contrastive Event Adjudication for Training-Free
Video Anomaly Detection} (CEAVAD).  CEAVAD first constructs hazard--benign event contrasts
from fixed public-safety knowledge.  Each contrast pairs a hazard mechanism with a generic
normal account, a mechanism-specific benign counterpart, and visible event-state evidence.  A frozen
vision--language encoder compares each target with the complete description banks and
composes their semantic contrasts into a target-specific, revisable boundary proposal.
Finally, a frozen MLLM adjudicates the competing explanations using
role-separated \textsc{Past}, \textsc{Target}, and \textsc{Future} observations.  The target
establishes event occurrence in the scored interval, while the surrounding episode tests
local normality and the adequacy of the benign explanation.  This adjudication yields
temporally localized anomaly scores and, from the same evidence, an evidence-grounded
explanation.

Our contributions are threefold:

\begin{itemize}
    \item We distinguish semantic evidence from the anomaly criterion and formulate
    training-free VAD through \emph{contrastive event adjudication} of falsifiable event
    hypotheses, establishing an inference-time explanatory boundary by proposing, testing,
    and revising competing interpretations against temporally organized video evidence.
    \item We develop CEAVAD, which structures public-safety knowledge as hazard--benign event
    contrasts, composes evidence from a frozen vision--language encoder into a composite and
    revisable contrastive boundary proposal, and adjudicates it with role-partitioned episodic
    evidence, yielding both temporally localized anomaly decisions and evidence-grounded
    explanations.
    \item The same fixed system reaches the state of the art among training-free methods on three widely used VAD benchmarks.
\end{itemize}

\section{Related Work}
\label{sec:related-work}

\subsection{Video Anomaly Detection}

Video anomaly detection is commonly studied under one-class or unsupervised learning and
weak supervision.  The former learns target-domain normality and uses reconstruction error
\cite{hasan2016learning}, future-frame prediction \cite{liu2018future}, or memory banks
\cite{park2020learning} as anomaly evidence.  The latter learns from video-level labels
through multiple-instance learning (MIL) ranking \cite{sultani2018real}, with subsequent
methods improving temporal
supervision through pseudo-label refinement \cite{feng2021mist}, feature-magnitude learning
\cite{tian2021rtfm}, or debiased MIL \cite{lv2023unbiased}.  VadCLIP incorporates transferable
vision--language features but still optimizes task-specific branches with weak labels
\cite{wu2024vadclip}.  Despite different supervision, these methods estimate the anomaly
criterion through optimization on VAD data.

\subsection{Training-Free Video Anomaly Detection}

Recent training-free VAD methods exploit frozen foundation models to avoid task-specific
detector optimization.  Building on the transferable semantics of vision--language
pretraining \cite{radford2021learning}, LAVAD generates frame captions, filters them using
cross-modal similarity, aggregates their temporal content with an LLM, and refines the
resulting anomaly estimates with vision--language similarity
\cite{zanella2024harnessing}.  MCANet extends this caption-based formulation to visual and
audio descriptions and consolidates their temporal semantics with an LLM
\cite{dev2024mcanet}.  AnomalyRuler instead induces scenario-specific normality rules from a
few normal reference videos and applies those rules deductively at test time
\cite{yang2024follow}.

Subsequent work improves the temporal organization of foundation-model inference.  EventVAD
models a dynamic spatiotemporal graph, detects event boundaries from unsupervised statistical
signals, and uses hierarchical prompting for event-level MLLM reasoning
\cite{shao2025eventvad}.  VADTree uses a pretrained generic event-boundary detector to build
an adaptive coarse-to-fine hierarchy, removes redundant nodes, and integrates
multi-granularity scores through inter-cluster node correlation \cite{li2025vadtree}.
Related tuning-free approaches extract anomaly cues from internal MLLM representations: HiProbe-VAD
selects informative intermediate hidden states for scoring, temporal localization, and
explanation \cite{cai2025hiprobe}, whereas HeadHunt-VAD derives detection features from a
sparse set of anomaly-sensitive attention heads \cite{cai2026headhunt}.

External knowledge further enriches training-free inference: DR-VAD uses real-world anomaly
definitions to guide structured temporal reasoning \cite{pei2026drvad}, while RAG4VAD
retrieves normal exemplars to refine a scene-aware normality baseline and explain deviations
\cite{sun2026rag4vad}.  CEAVAD instead distinguishes evidence from the anomaly criterion,
treating competing explanations as falsifiable event hypotheses and adjudicating a revisable
contrastive boundary proposal against temporally organized video evidence.

\begin{figure*}[t]
    \centering
    \includegraphics[width=\textwidth]{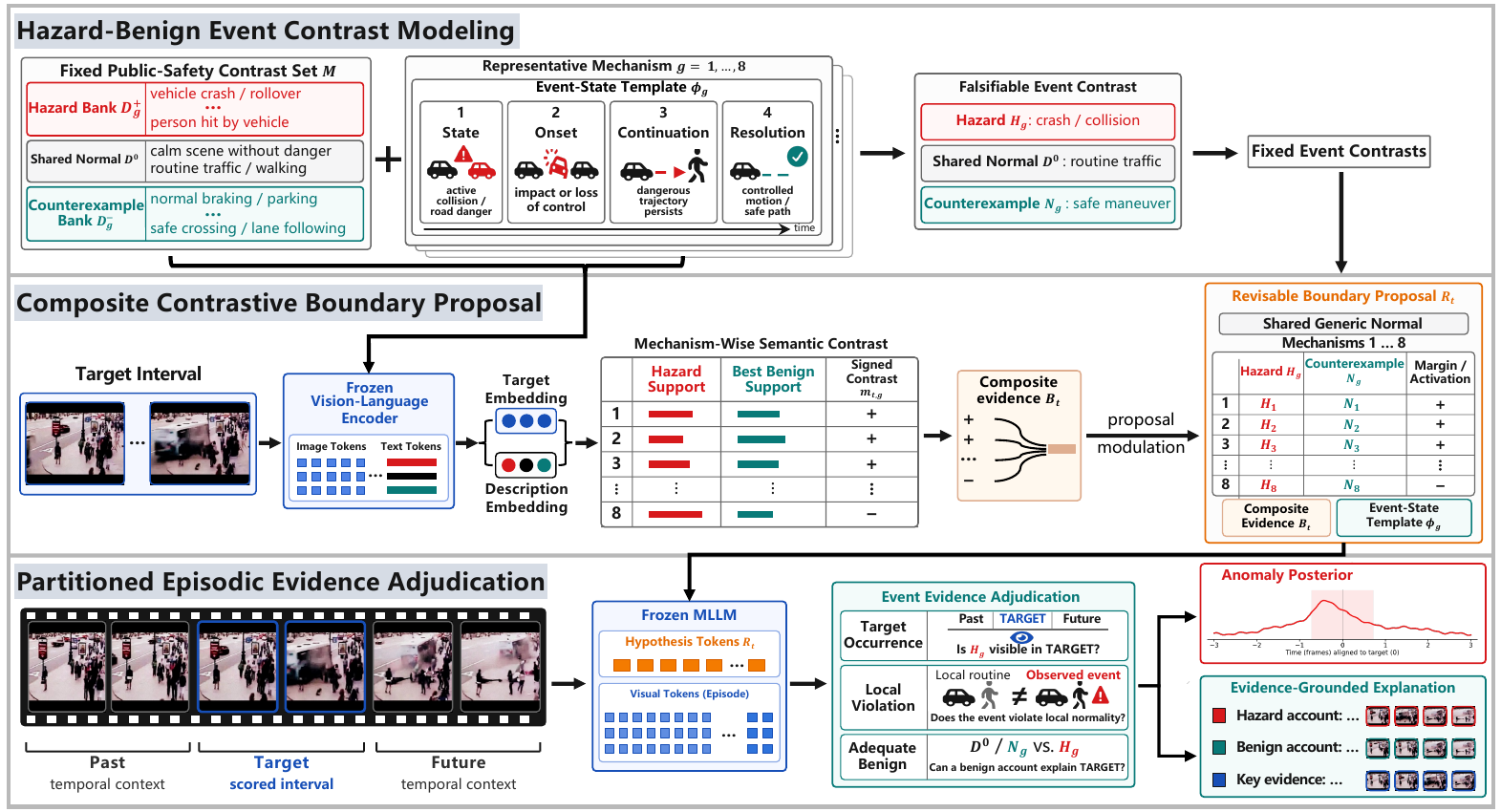}
    \caption{Overview of the CEAVAD framework.  First, hazard--benign event contrast modeling
    structures fixed public-safety knowledge into falsifiable mechanism-level hypotheses with
    event-state templates.  Next, a frozen vision--language encoder computes mechanism-wise
    semantic contrasts and conditions their proposal margins on a composite semantic proposal
    score, producing a revisable boundary proposal.  Finally, a frozen MLLM adjudicates the
    proposal with role-separated \textsc{Past}, \textsc{Target}, and \textsc{Future}
    observations, yielding the two-token anomaly score and an evidence-grounded
    explanation.}
    \label{fig:framework}
\end{figure*}

\section{Method}
\label{sec:method}

Video anomaly detection (VAD) identifies abnormal events and localizes them on the video
timeline.  Given a video $V=\{I_f\}_{f=1}^{F}$, we divide it into target intervals
$\{W_t\}_{t=1}^{T}$, assign each interval an anomaly score $p_t\in[0,1]$, and map these scores
to a frame-aligned sequence $\mathbf{s}(V)=(s_1,\ldots,s_F)$.  We propose \emph{Contrastive
Event Adjudication for Training-Free Video Anomaly Detection} (CEAVAD), which performs this
inference with frozen pretrained models and no target-domain parameter fitting.  CEAVAD
casts each target as a competition between falsifiable hazard and benign event hypotheses,
establishing an inference-time \emph{explanatory boundary} through their interaction with
video evidence.  Target-level semantic contrast forms a revisable boundary proposal, which
CEAVAD then adjudicates against temporally organized video evidence.

The framework comprises three stages.  First, CEAVAD organizes fixed public-safety knowledge
into hazard--benign event contrasts, pairing each hazard with a generic normal account and a
mechanism-specific benign alternative (Sec.~\ref{sec:event_contrasts}).  Second, a frozen
vision--language encoder contrasts the target against these explanations and composes their
evidence into a revisable boundary proposal (Sec.~\ref{sec:boundary_proposal}).  Third, a
frozen MLLM adjudicates the proposal with role-separated \textsc{Past}, \textsc{Target}, and
\textsc{Future} evidence, yielding the two-token anomaly score
and, on demand, an evidence-grounded explanation
(Sec.~\ref{sec:episodic_adjudication}).  Figure~\ref{fig:framework} illustrates the complete
flow and the information passed between these stages.

\subsection{Hazard--Benign Event Contrast Modeling}
\label{sec:event_contrasts}

CEAVAD first represents public-safety knowledge as a fixed contrast set
$\mathcal{M}=\{\mathcal{M}_g\}_{g=1}^{G}$ with $G=8$.  Each mechanism is modeled as
\begin{equation}
\mathcal{M}_g=
\bigl(\mathcal{D}^{+}_g,\mathcal{D}^{0},\mathcal{D}^{-}_g,\Phi_g\bigr),
\label{eq:contrast_unit}
\end{equation}
where $\mathcal{D}^{+}_g$ describes observable realizations of the hazard,
$\mathcal{D}^{-}_g$ describes visually similar nonhazardous counterexamples, and the shared bank
$\mathcal{D}^{0}$ describes ordinary activity.  Each mechanism contains three hazard and
three mechanism-specific counterexample descriptions, while $\mathcal{D}^{0}$ contains two
generic normal descriptions.  These banks form a falsifiable contrast between the hazard
hypothesis and two complementary alternatives: ordinary activity and a visually confusable,
mechanism-specific counterexample; see the supplement.

CEAVAD uses these description banks in two successive inference stages.  First, the frozen
vision--language encoder embeds every description in
$\mathcal{D}^{+}_g$, $\mathcal{D}^{-}_g$, and $\mathcal{D}^{0}$ and aggregates their
similarities into the set-level support defined in Sec.~\ref{sec:boundary_proposal}.  Using
multiple descriptions at this stage covers different visual realizations of the same event
mechanism.  The MLLM then adjudicates one explicit competition per mechanism in
Sec.~\ref{sec:episodic_adjudication}: a fixed hazard statement
$H_g\in\mathcal{D}^{+}_g$ against a fixed mechanism-specific counterexample
$N_g\in\mathcal{D}^{-}_g$, together with the shared normal references and the target-specific
margin and activation computed from the complete description banks.
The remaining descriptions contribute through these target-specific quantities.  This
retains their retrieval coverage without presenting paraphrases of the same mechanism as
separate event hypotheses to the MLLM.

To make each competition testable over time, the remaining component $\Phi_g$ specifies the
required visible evidence:
\begin{equation}
\Phi_g=\bigl(\phi_g^{\mathrm{state}},\phi_g^{\mathrm{onset}},
\phi_g^{\mathrm{cont}},\phi_g^{\mathrm{end}}\bigr).
\label{eq:event_state_template}
\end{equation}
Here, $\phi_g^{\mathrm{state}}$ specifies the unsafe relation required in the target;
$\phi_g^{\mathrm{onset}}$, $\phi_g^{\mathrm{cont}}$, and $\phi_g^{\mathrm{end}}$ describe its
onset, continuation, and resolution through termination or benign evidence.  For example, a
vehicle-hazard template uses impact or loss of control as onset, a dangerous trajectory or
wreckage as continuation, and controlled braking or parking as benign resolution.  The MLLM
checks the target for the unsafe state and uses ordered context to distinguish onset or
continuation from resolution.  Thus $\Phi_g$ makes each contrast testable, supporting the hazard
hypothesis or the corresponding benign account according to visible evidence.

\subsection{Composite Contrastive Boundary Proposal}
\label{sec:boundary_proposal}

CEAVAD next instantiates the fixed contrasts for each target.  Let
$X_t^{0}=\{I_{t,j}\}_{j=1}^{n_t}$ denote the frames sampled from $W_t$.  The frozen
vision--language encoder maps each frame to a unit vector
$\mathbf{v}_{t,j}$; their mean is normalized once more to obtain the target representation
\begin{equation}
\mathbf{q}_t=
\frac{\frac{1}{n_t}\sum_{j=1}^{n_t}\mathbf{v}_{t,j}}
{\left\|\frac{1}{n_t}\sum_{j=1}^{n_t}\mathbf{v}_{t,j}\right\|_2}.
\label{eq:target_embedding}
\end{equation}
Only target frames contribute to $\mathbf{q}_t$.  In parallel, the frozen text encoder maps
each description to a unit vector.  We write $\mathcal{P}^{+}_g$, $\mathcal{P}^{-}_g$, and
$\mathcal{P}^{0}$ for the resulting embedding sets corresponding to
$\mathcal{D}^{+}_g$, $\mathcal{D}^{-}_g$, and $\mathcal{D}^{0}$, respectively.

With the target and description embeddings in place, CEAVAD aggregates the target's semantic
support for any text-embedding set $\mathcal{P}$ over its complete description bank as
\begin{equation}
E(\mathbf{q}_t,\mathcal{P})=
\log\left[
\frac{1}{|\mathcal{P}|}
\sum_{\mathbf{p}\in\mathcal{P}}
\exp(\mathbf{q}_t^{\top}\mathbf{p})
\right].
\label{eq:set_support}
\end{equation}
The log-mean-exp accumulates support across the bank while keeping a strongly matched
description influential.  On this basis, CEAVAD measures the signed contrast for mechanism
$g$ against the more strongly supported benign account:
\begin{equation}
m_{t,g}=E(\mathbf{q}_t,\mathcal{P}^{+}_g)
-\max\!\left\{
E(\mathbf{q}_t,\mathcal{P}^{0}),
E(\mathbf{q}_t,\mathcal{P}^{-}_g)
\right\}.
\label{eq:mechanism_contrast}
\end{equation}
A positive $m_{t,g}$ means that the target semantically supports the hazard bank more than
both the generic normal account and its mechanism-specific benign counterpart.  Its sign is
only a provisional comparison of textual explanations; event occurrence has not yet been
established.

These signed margins are raw cosine gaps.  Using them for all targets underuses the
contrastive precision learned by the frozen encoder, whereas applying that precision
indiscriminately can amplify an isolated positive match even when the complete hazard family
is unsupported.  CEAVAD therefore conditions the use of the model-native scale on a composite
semantic proposal score.  Let
$\gamma=\exp(\ell_{\mathrm{enc}})$ be the encoder's frozen logit scale.  We compute
\begin{equation}
B_t=
\log\left[
\frac{1}{G}\sum_{g=1}^{G}
\exp\!\left(\gamma m_{t,g}\right)
\right].
\label{eq:composite_evidence}
\end{equation}
$B_t$ is the composite semantic proposal score, summarizing family-level contrastive evidence
without selecting a mechanism.  Because it is a log-mean-exp, the evidence can be broad or
concentrated in one well-supported
mechanism; it is not a vote among mechanisms.  Zero is the fixed neutral reference inherited
from the signed hazard--benign contrasts.  The proposal margin is then
\begin{equation}
\widetilde m_{t,g}=
\left\{
\begin{array}{ll}
\gamma m_{t,g}, & B_t>0\ \mathrm{and}\ m_{t,g}>0,\\
m_{t,g}, & \mathrm{otherwise},
\end{array}
\right.
\label{eq:proposal_state}
\end{equation}
We expose this proposal margin together with its bounded activation
$a_{t,g}=1/[1+\exp(-\widetilde m_{t,g})]$.
When $B_t>0$, the native scale sharpens only the
positively supported mechanisms.  Otherwise every mechanism retains its raw contrast, and a
nonpositive contrast is never turned positive by the adjustment.  The sign test and scale
come entirely from the fixed formulation and frozen encoder; they are shared by all target
videos.

CEAVAD then assembles these quantities into the complete boundary proposal
$\mathcal{R}_t$, which contains the generic normal account.  For
every mechanism, it also contains $H_g$, $N_g$, the state description $\Phi_g$, the proposal
margin $\widetilde m_{t,g}$, and activation $a_{t,g}$.  When $B_t>0$,
the proposal retains $B_t$ and the raw margins as well.  No mechanism is pruned because of
its margin or rank.  Thus $\mathcal{R}_t$ expresses which side of each hazard--benign
contrast the target currently supports while keeping every explanation available for visual
revision.  By coupling mechanism-level semantic support with the corresponding event
hypotheses and state descriptions, $\mathcal{R}_t$ bridges target-level retrieval and
video-grounded adjudication.

\subsection{Partitioned Episodic Evidence Adjudication}
\label{sec:episodic_adjudication}

CEAVAD finally adjudicates $\mathcal{R}_t$ against temporally ordered video evidence.  The
proposal summarizes the target's relative semantic support for the competing event
explanations, while a target-centered episode reveals how the event unfolds around the
target.  To this end, CEAVAD organizes the episode into \textsc{Past}, \textsc{Target}, and
\textsc{Future} observations.

We divide the video into nonoverlapping two-second intervals, sample one frame per second,
and use the immediately adjacent intervals as context for $W_t$.  The final interval is
truncated at the video end; sampling is mapped to valid frames and unavailable context is
omitted.  The ordered groups $\widetilde X_t^{-}$, $\widetilde X_t^{0}$, and
$\widetilde X_t^{+}$ contain up to two \textsc{Past}, \textsc{Target}, and \textsc{Future}
observations, respectively.

CEAVAD then conditions the MLLM on the resulting episode and proposal for $W_t$:
\begin{equation}
\mathcal{C}_t=
\bigl(\widetilde X_t^{-},\widetilde X_t^{0},
\widetilde X_t^{+},\mathcal{R}_t\bigr).
\label{eq:adjudication_condition}
\end{equation}
The three image groups are explicitly labeled \textsc{Past Context}, \textsc{Target Segment},
and \textsc{Future Context}; the target is marked as the sole interval being scored.  The MLLM
receives all eight hazard--benign contrasts, their proposal margins and activations, and their
event-state descriptions.  It compares these candidate explanations with visible people,
objects, actions, relations, and consequences in the episode.

Within $\mathcal{C}_t$, the temporal groups assume different inferential roles.  The
\textsc{Target} channel must establish that the concrete event described by some $H_g$
actually occurs in $W_t$.  Once that occurrence is visible, \textsc{Past} and \textsc{Future}
help assess whether it violates local normality or is adequately explained by the generic
normal account or $N_g$.  The event-state description $\Phi_g$ connects evidence of onset,
continuation, and resolution across these groups.  The necessary evidential conditions for
accepting a hazard hypothesis can be summarized as
\begin{equation}
\begin{array}{c}
\mathrm{Accept}_t(H_g)\ \Longrightarrow\\[2pt]
\mathrm{TargetOccurrence}_t(H_g)\\[2pt]
{}\land\ \mathrm{LocalViolation}_t(H_g)\\[2pt]
{}\land\ \neg\mathrm{AdequateBenign}_t(N_g,\mathcal{D}^{0}).
\end{array}
\label{eq:adjudication_criterion}
\end{equation}
Target evidence is required to establish event occurrence.  Context cannot substitute for
this evidence, but it informs whether the target-visible event violates local normality or is
adequately explained by a benign alternative.  A hazard confined to an adjacent group is not
transferred to $W_t$.  The MLLM may therefore confirm a target-visible hazard, reject it in
favor of a benign explanation, or treat insufficient target evidence as normal.

Let $z_{t,a}$ and $z_{t,n}$ be the language-head logits under $\mathcal{C}_t$ for the
leading-space, single-token continuations \texttt{abnormal} and \texttt{normal}.  Their
normalized two-token posterior is
\begin{equation}
p_t=
\frac{\exp(z_{t,a})}
{\exp(z_{t,a})+\exp(z_{t,n})}.
\label{eq:final_posterior}
\end{equation}
This posterior is the interval-level anomaly score, with $\mathcal{R}_t$ influencing it through
$\mathcal{C}_t$.  Since the target intervals are nonoverlapping, each frame inherits its unique
interval's score:
\begin{equation}
s_f=p_t,\qquad I_f\in W_t.
\label{eq:frame_mapping}
\end{equation}

Beyond this score, the same condition supports an evidence-grounded explanation.  The
MLLM reports target-visible evidence, temporal state, and how the episode resolves competing
explanations.  Shared evidence supports localized detection and explanation.

\section{Experiments}
\label{sec:experiments}

\subsection{Experimental Setup}
\label{sec:experimental_setup}

\paragraph{Datasets and metrics.}
We evaluate on three widely used benchmarks with distinct visual domains.  UCF-Crime
\cite{sultani2018real} contains long real-world surveillance videos; its test set comprises
290 videos.  UBnormal \cite{acsintoae2022ubnormal} contains 211 test videos rendered across
multiple virtual scenes, with anomaly types disjoint between training and testing.
XD-Violence \cite{wu2020notonly} contains diverse real-world and cinematic violence; we
evaluate all 800 test videos using the visual stream.  Following standard protocols, we
report frame-level ROC-AUC for UCF-Crime, micro frame-level ROC-AUC for UBnormal, and
both frame-level ROC-AUC and average precision (AP) for XD-Violence, where AP is the primary
metric.  All metrics are computed over the complete test sets.

\paragraph{Unified training-free protocol.}
CEAVAD is training-free: it performs no parameter updates or target-domain fitting, and all
pretrained parameters remain frozen.  It is zero-shot: no target-domain demonstrations, labels,
or retrieved references are provided at inference.
All three datasets share the same eight-mechanism public-safety vocabulary,
the same model configuration, and the same scoring rule.  Videos are divided into
nonoverlapping two-second target intervals and sampled at one frame per second; the adjacent
intervals provide the \textsc{Past} and \textsc{Future} observations.  The frozen
vision--language encoder is OpenCLIP ViT-L/14 with MetaCLIP fullcc weights, and the frozen
MLLM is Qwen3-VL-8B.  The reported score is the 32-bit floating-point (FP32) two-token
posterior over the continuations \texttt{abnormal} and \texttt{normal}, as defined in
Eq.~\ref{eq:final_posterior}.  All experiments are executed on a single NVIDIA H20 GPU.

\begin{table*}[t]
\centering
\small
\setlength{\tabcolsep}{3.8pt}
\begin{tabular*}{\textwidth}{@{\extracolsep{\fill}}lcccccc@{}}
\toprule
& & & \multicolumn{1}{c}{UCF-Crime} & \multicolumn{1}{c}{UBnormal} & \multicolumn{2}{c}{XD-Violence} \\
\cmidrule(lr){4-5}\cmidrule(lr){6-7}
Method & Zero-shot & Training-free & \multicolumn{1}{c}{ROC-AUC (\%)} & \multicolumn{1}{c}{ROC-AUC (\%)} & \multicolumn{1}{c}{ROC-AUC (\%)} & \multicolumn{1}{c}{AP (\%)} \\
\midrule
\multicolumn{7}{l}{Training-dependent methods} \\
RTFM \cite{tian2021rtfm} & No & No & 84.30 & 64.94 & -- & 77.81 \\
UR-DMU \cite{zhou2023dual} & No & No & 86.97 & 59.91 & -- & 81.66 \\
VadCLIP \cite{wu2024vadclip} & No & No & 88.02 & -- & -- & 84.51 \\
Holmes-VAU \cite{zhang2025holmesvau} & Yes & No & 88.96 & 56.77 & -- & 87.68 \\
\midrule
\multicolumn{7}{l}{Training-free methods} \\
AnomalyRuler \cite{yang2024follow} & No & Yes & -- & 71.90 & -- & -- \\
LAVAD \cite{zanella2024harnessing} & Yes & Yes & 80.28 & 64.23 & 85.36 & 62.01 \\
EventVAD \cite{shao2025eventvad} & Yes & Yes & 82.03 & -- & 87.51 & 64.04 \\
SUVAD \cite{gao2025suvad} & Yes & Yes & 83.90 & -- & -- & 70.10 \\
MCANet \cite{dev2024mcanet} & Yes & Yes & 82.47 & -- & 87.43 & 69.72 \\
VADTree \cite{li2025vadtree} & Yes & Yes & 84.74 & -- & 90.44 & 67.82 \\
PANDA \cite{yang2025panda} & Yes & Yes & 84.89 & 75.78 & -- & 70.16 \\
IntraTR (adaptive) \cite{lin2025unified} & Yes & Yes & 84.08 & 69.02 & 91.23 & 68.03 \\
CoReVAD \cite{lim2026corevad} & Yes & Yes & 82.51 & -- & 91.44 & 70.94 \\
\midrule
\textbf{CEAVAD} & \textbf{Yes} & \textbf{Yes} & \textbf{86.15} & \textbf{77.55} & \textbf{93.40} & \textbf{79.52} \\
\bottomrule
\end{tabular*}
\caption{Comparison on three VAD benchmarks.  Bold marks the best zero-shot,
training-free result.}
\label{tab:main_results}
\end{table*}

\subsection{Main Results}
\label{sec:main_results}

Table~\ref{tab:main_results} compares CEAVAD with training-dependent methods and with methods
that perform inference without parameter updates, while distinguishing whether target-domain
references are used.  CEAVAD achieves 86.15\% ROC-AUC on UCF-Crime and 77.55\% micro
ROC-AUC on UBnormal, improving the previous best zero-shot, training-free results by 1.26 and
1.77 percentage points, respectively.  On XD-Violence, it reaches 93.40\% ROC-AUC and 79.52\%
AP, exceeding the strongest prior zero-shot, training-free results by 1.96 and 8.58 points.
It also surpasses RTFM across the three primary metrics, although some methods trained with
weak labels or anomaly-oriented instructions retain higher UCF-Crime ROC-AUC and XD-Violence AP.

\subsection{Ablation Studies}
\label{sec:ablations}

Table~\ref{tab:component_ablations} compares CEAVAD with Direct Qwen and three targeted
variants.  Direct Qwen removes the event contrasts and semantic proposal; the variants remove
the event-state descriptions $\Phi_g$, retain only the highest-margin mechanism for visual
adjudication, or remove the distinct evidential roles of \textsc{Past}, \textsc{Target}, and
\textsc{Future}, respectively.  Full CEAVAD improves over Direct Qwen by 5.55 and 7.14
ROC-AUC points on UCF-Crime and UBnormal, respectively.  Across the two datasets, removing
event-state descriptions costs 1.87 and 3.54 points; retaining only the top mechanism costs
2.73 and 6.12 points; and removing the evidence partition costs 1.04 and 1.97 points.  These
consistent drops support event-state cues, cross-mechanism visual revision, and role-specific
temporal evidence as complementary components.  Additional variants are reported in the
supplementary material.

\begin{table}[t]
\centering
\small
\begin{tabular}{@{}ccc@{}}
\toprule
\makebox[0.40\columnwidth][c]{Method} &
\makebox[0.245\columnwidth][c]{\shortstack{UCF\\ROC-AUC (\%)}} &
\makebox[0.245\columnwidth][c]{\shortstack{UB\\ROC-AUC (\%)}} \\
\midrule
Direct Qwen & 80.60 & 70.41 \\
w/o event-state & 84.28 & 74.01 \\
Top-1 mechanism & 83.42 & 71.43 \\
w/o evidence partition & 85.11 & 75.58 \\
\textbf{Full CEAVAD} & \textbf{86.15} & \textbf{77.55} \\
\bottomrule
\end{tabular}
\caption{Key design ablations on UCF-Crime and UBnormal.}
\label{tab:component_ablations}
\end{table}

\subsection{Explanatory-Boundary Analysis}
\label{sec:boundary_analysis}

\begin{figure}[t]
\centering
\includegraphics[width=\columnwidth]{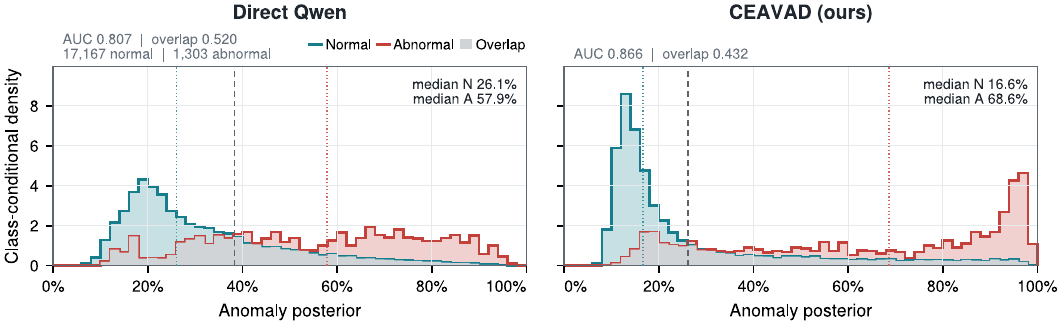}
\caption{Distributional analysis of CEAVAD's inference-time explanatory boundary.}
\label{fig:posterior_distribution}
\end{figure}

\paragraph{Posterior separation.}
We first examine whether CEAVAD produces a more discriminative organization of anomaly
evidence at the dataset level.  Figure~\ref{fig:posterior_distribution} compares the
class-conditional posterior distributions of Direct Qwen and CEAVAD on all unambiguous
UCF-Crime target intervals: 17,167 normal and 1,303 abnormal intervals.  We exclude 198
transition intervals containing both normal and abnormal content, and select no interval
according to either method's posterior.  The left and right panels show Direct Qwen and
CEAVAD, respectively.  Cyan and red curves are class-normalized densities for normal and
abnormal intervals, the gray region is their intersection, colored dotted lines mark class
medians, and the black dashed line marks only the density crossover rather than a prediction
threshold.

\begin{figure}[t]
\centering
\includegraphics[width=\columnwidth]{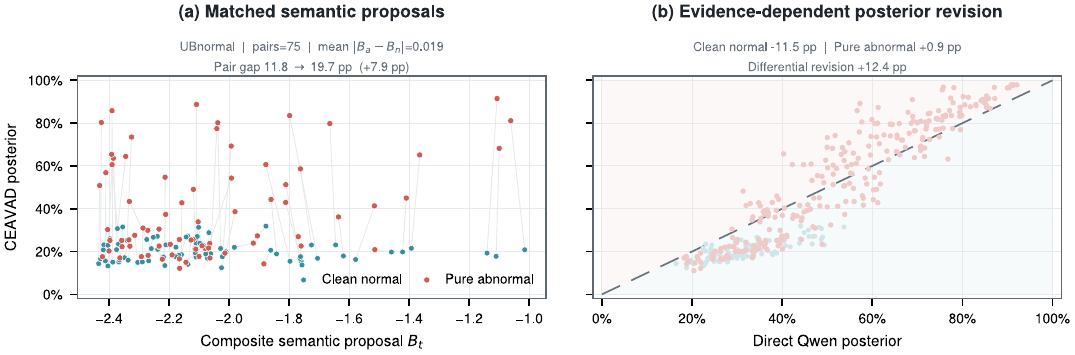}
\caption{Matched proposals and evidence-dependent posterior revision on UBnormal.}
\label{fig:ubnormal_boundary}
\end{figure}

\begin{figure}[b]
\centering
\includegraphics[width=\columnwidth]{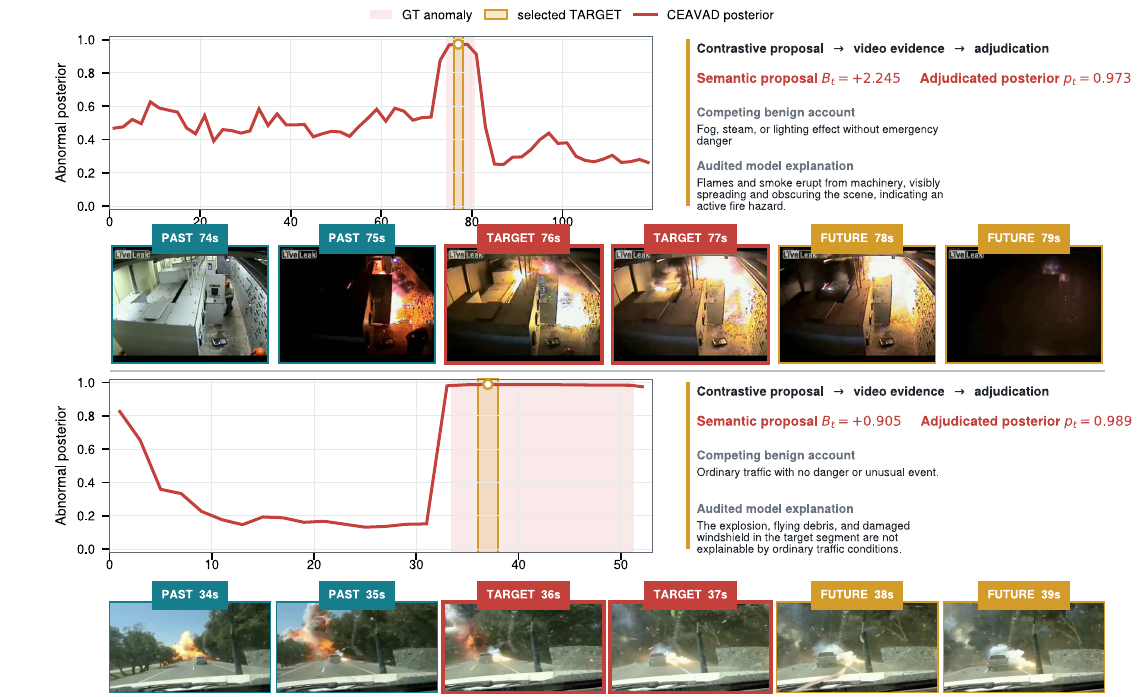}
\caption{End-to-end visualization of contrastive proposal and evidence-grounded adjudication in CEAVAD.}
\label{fig:full_chain}
\end{figure}

Direct Qwen already provides a useful ranking signal, but its normal and abnormal posteriors
remain substantially entangled: their medians are 26.1\% and 57.9\%, giving a separation of
31.8 percentage points and an overlap coefficient of 0.520.  CEAVAD moves the medians to
16.6\% and 68.6\%, increasing the separation to 52.0 points and reducing the overlap to
0.432.  The diagnostic interval-level AUC on this unambiguous subset rises from 0.807 to
0.866; this diagnostic is used only to quantify the plotted distributions and is distinct
from the official benchmark metrics in Table~\ref{tab:main_results}.  The opposing median
shifts and smaller intersection show that CEAVAD changes the organization of normal and
abnormal evidence in a class-dependent way, rather than applying a shared posterior offset.

\paragraph{Posterior-blind diagnostic.}
The aggregate distributions can still differ because abnormal intervals may begin with
stronger semantic hazard support.  To test for evidence-dependent revision after controlling
for the top mechanism and $B_t$, the composite semantic proposal score, we construct a
posterior-blind matched-pair diagnostic.  A target is labeled \emph{pure abnormal} when at
least 95\% of its frames are annotated abnormal; it is \emph{clean normal} when neither the
target nor its adjacent context overlaps an annotated anomaly.  Transition targets are
excluded.  Within each mechanism, we match a pure-abnormal target to a clean-normal target
from a different video without replacement, requiring $|B_a-B_n|$ to be at most 0.1 pooled
standard deviations.  UCF-Crime and
XD-Violence use $B_t>0$ to form the high-resemblance pool; on UBnormal this rule yields no
clean-normal candidate, so we use the upper quartile of the $B_t$ distribution.  These
selection and matching operations use only labels and $B_t$; neither posterior is used for
selection or matching.  Confidence intervals are obtained
from 2,000 video-level clustered bootstrap replicates, keeping temporally adjacent targets
from the same video in one resampling cluster.

Let $\mathcal{H}_1$ and $\mathcal{H}_0$ denote the pure-abnormal and clean-normal targets in
the pre-matching high-resemblance population, and let $\mathcal{P}\subseteq
\mathcal{H}_1\times\mathcal{H}_0$ denote the matched pairs.  For CEAVAD and Direct Qwen
posteriors $p_t^{\mathrm{C}}$ and $p_t^{\mathrm{D}}$, respectively, the mean matched gap gain is
\begin{equation}
\widehat{G}_{\mathcal{P}}=\frac{1}{|\mathcal{P}|}
\sum_{(a,n)\in\mathcal{P}}
\left[\bigl(p_a^{\mathrm{C}}-p_n^{\mathrm{C}}\bigr)-
\bigl(p_a^{\mathrm{D}}-p_n^{\mathrm{D}}\bigr)\right].
\label{eq:gap_gain}
\end{equation}
For the posterior revision $\delta_t=p_t^{\mathrm{C}}-p_t^{\mathrm{D}}$, we separately use
the complete pre-matching population to compute
\begin{equation}
\Delta_{\mathcal{H}}=
\frac{1}{|\mathcal{H}_1|}\sum_{a\in\mathcal{H}_1}\delta_a-
\frac{1}{|\mathcal{H}_0|}\sum_{n\in\mathcal{H}_0}\delta_n.
\label{eq:differential_revision}
\end{equation}
The first statistic measures additional separation among matched targets; the second retains
all eligible targets and removes revision shared by the two classes.  Because they use
different populations, their estimates need not coincide.  Table~\ref{tab:boundary_diagnostics}
reports posterior percentage points with 95\% CIs from the clustered bootstrap.

\begin{table}[t]
\centering
\small
\setlength{\tabcolsep}{1.15pt}
\begin{tabular*}{\columnwidth}{@{\extracolsep{\fill}}lrrrcc@{}}
\toprule
Dataset & $|\mathcal{P}|$ & $|\mathcal{H}_0|$ & $|\mathcal{H}_1|$ &
$\widehat{G}_{\mathcal{P}}$ & $\Delta_{\mathcal{H}}$ \\
\midrule
UCF-Crime & 992 & 4,337 & 1,049 &
\shortstack[c]{$2.77$\\$[0.82,4.86]$} &
\shortstack[c]{$5.32$\\$[2.41,8.13]$} \\
UBnormal & 75 & 96 & 264 &
\shortstack[c]{$7.89$\\$[2.78,13.32]$} &
\shortstack[c]{$12.40$\\$[9.42,15.60]$} \\
XD-Violence & 3,599 & 5,593 & 8,127 &
\shortstack[c]{$1.17$\\$[-0.12,2.44]$} &
\shortstack[c]{$2.31$\\$[1.19,3.45]$} \\
\bottomrule
\end{tabular*}
\caption{Posterior-blind matched-pair and pool-level diagnostics.}
\label{tab:boundary_diagnostics}
\end{table}

The proposal-score matching is tight: mean $|B_a-B_n|$ is 0.0053, 0.0190, and 0.0097 on
UCF-Crime, UBnormal, and XD-Violence, respectively.  Under this controlled semantic
resemblance, the abnormal--normal pair gap grows by 2.77 points on UCF-Crime and 7.89 points
on UBnormal; the XD-Violence estimate is positive, although its interval crosses zero.  The
pool-level class-differential revision is positive on all three datasets, with each 95\% CI
excluding zero.  Together, the diagnostics provide evidence consistent with a systematic difference
in how CEAVAD revises abnormal and normal targets; the matched estimate controls the top
mechanism and $B_t$, while the pool-level estimate does not require every normal score to
decrease.  UCF-Crime and XD-Violence also contain shared upward shifts.

\paragraph{Evidence-dependent boundary revision.}
Figure~\ref{fig:ubnormal_boundary} visualizes this revision on UBnormal using the matched
pairs defined above.  Panel (a) displays all 75 matched pairs; gray segments connect the
endpoints, with cyan and red denoting clean-normal and pure-abnormal targets.  The mean
absolute proposal-score difference is only 0.019, and the standardized difference between the two
proposal-score distributions is $-0.006$.  Direct Qwen produces an 11.8-point mean
abnormal--normal gap, whereas CEAVAD produces 19.7 points, a gain of 7.9 points with a 95\%
CI of $[2.8,13.3]$.  Pairwise ranking accuracy increases from 65.3\% to 80.0\%.

\FloatBarrier

Panel (b) explains where this separation comes from.  We use the pre-specified high-resemblance
subset whose $B_t$ is at or above the dataset-wide upper quartile.  The horizontal axis is the
Direct Qwen posterior, the vertical axis is the CEAVAD posterior, and the dashed diagonal
denotes an unchanged posterior.  Among 96 high-resemblance clean-normal intervals, CEAVAD
changes the posterior by $-11.5$ points on average (95\% CI $[-13.5,-9.2]$); among 264
pure-abnormal intervals, the mean change is $+0.9$ points.  The resulting class-differential
revision is $+12.4$ points (95\% CI $[9.4,15.6]$).  The matched-pair and revision views
therefore provide complementary evidence: the former controls the top mechanism and $B_t$,
whereas the latter measures class-dependent posterior revision over all high-resemblance
targets.

\paragraph{End-to-end qualitative analysis.}
Figure~\ref{fig:full_chain} visualizes the end-to-end CEAVAD process across diverse benchmark
videos.  Each row connects the temporally localized anomaly response with the
contrastive proposal, video-evidence adjudication, and evidence-grounded explanation.
Together, the examples illustrate how the same adjudication process supports both temporally
localized anomaly detection and evidence-grounded anomaly understanding.

\FloatBarrier

\section{Conclusion}
\label{sec:conclusion}

We introduced CEAVAD, a training-free VAD method that turns pretrained semantics into an
inference-time anomaly criterion through contrastive, evidence-grounded event adjudication.
CEAVAD constructs falsifiable hazard--benign event contrasts, forms a revisable proposal from
target-level semantic support, and adjudicates competing explanations with temporally structured
video evidence.  The resulting explanatory boundary supports localized anomaly detection and
evidence-grounded explanations.  The same fixed system sets the training-free state of the art
on UCF-Crime, UBnormal, and XD-Violence, with ablations and boundary diagnostics further
supporting the design.

\bibliography{references}

\end{document}


\makesupplementtitle

\section{Supplementary Overview}
\label{sec:supp_overview}

This supplement provides the complete specification and extended evaluation of CEAVAD.  It
first expands the method details condensed in the main paper, including the fixed public-safety
vocabulary, event-state descriptions, prompt condition, score construction, and evaluation
mapping.  It then reports extended component ablations, temporal and textual sensitivity
analyses, and additional qualitative cases.
All benchmark results use the same training-free protocol and the same fixed
eight-mechanism hypothesis space described in the main paper.

\section{Complete Method Specification}
\label{sec:supp_method}

This section specifies the complete path from an input video to its frame-aligned anomaly
posterior.  The notation follows Sec.~3 of the main paper.  Given
$V=\{I_f\}_{f=1}^{F}$, CEAVAD constructs nonoverlapping target intervals
$\{W_t\}_{t=1}^{T}$, forms one revisable boundary proposal $\mathcal{R}_t$ for each target,
and adjudicates that proposal with a target-centered video episode.  No target-domain
parameter is estimated in this process.

\begin{figure*}[t]
\centering
\includegraphics[width=\textwidth]{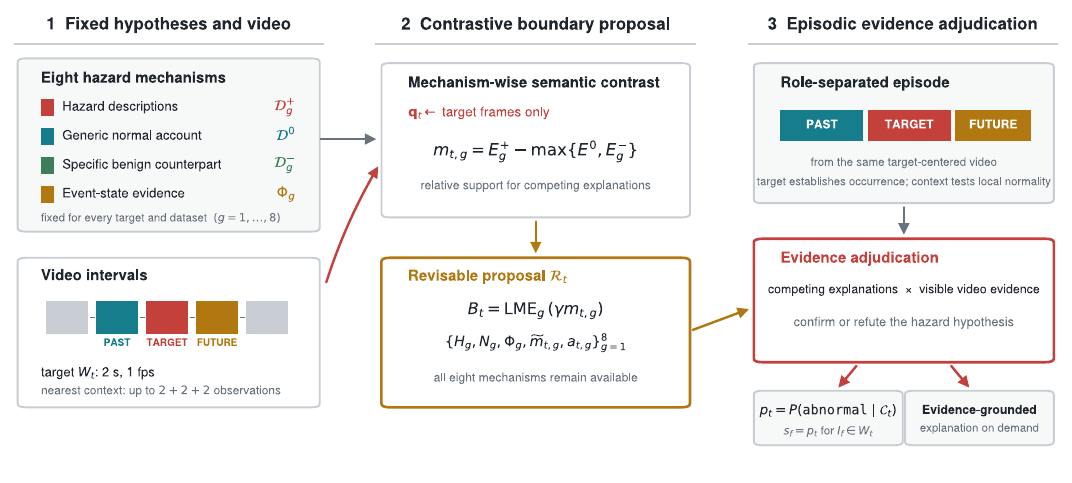}
\caption{Complete CEAVAD inference data flow.}
\label{fig:supp_inference_flow}
\end{figure*}

Figure~\ref{fig:supp_inference_flow} summarizes the complete data flow.  The target frames
instantiate the fixed event contrasts and form $\mathcal{R}_t$, while the target-centered
episode supplies the role-separated evidence used to adjudicate that proposal.
Algorithm~\ref{alg:ceavad} gives the corresponding per-video procedure.

\subsection{Fixed Event Contrasts}

The public-safety hypothesis space is
$\mathcal{M}=\{\mathcal{M}_g\}_{g=1}^{G}$ with $G=8$ and
\begin{equation}
\mathcal{M}_g=(\mathcal{D}^{+}_g,\mathcal{D}^{0},
\mathcal{D}^{-}_g,\Phi_g).
\label{eq:supp_contrast_unit}
\end{equation}
The hazard bank $\mathcal{D}^{+}_g$ and mechanism-specific benign bank
$\mathcal{D}^{-}_g$ each contain three observable descriptions.  The two-description bank
$\mathcal{D}^{0}$ is shared across mechanisms and represents generic normal activity.  Each
event-state specification
$\Phi_g=(\phi_g^{\mathrm{state}},\phi_g^{\mathrm{onset}},
\phi_g^{\mathrm{cont}},\phi_g^{\mathrm{end}})$ states the unsafe relation and the visible
evidence for its onset, continuation, and termination or benign resolution.  The complete
vocabulary is reported in Sec.~\ref{sec:supp_vocabulary}.

The description banks and the explicit MLLM hypotheses use the same event mechanisms at two
different resolutions.  All descriptions in $\mathcal{D}^{+}_g$, $\mathcal{D}^{-}_g$, and
$\mathcal{D}^{0}$ participate in visual--language retrieval.  The MLLM receives one fixed
canonical hazard statement $H_g\in\mathcal{D}^{+}_g$ and one canonical mechanism-specific
benign statement $N_g\in\mathcal{D}^{-}_g$ per mechanism.  Evidence from the complete banks
remains present through the target-specific margin and activation supplied with this
canonical competition.

\subsection{Target-Specific Boundary Proposal}

For the sampled target frames $X_t^0=\{I_{t,j}\}_{j=1}^{n_t}$, the frozen vision--language
encoder produces unit image embeddings $\mathbf{v}_{t,j}$.  Their normalized mean is
\begin{equation}
\overline{\mathbf{v}}_t=n_t^{-1}\sum_{j=1}^{n_t}\mathbf{v}_{t,j},\qquad
\mathbf{q}_t=
\frac{\overline{\mathbf{v}}_t}
{\max(\|\overline{\mathbf{v}}_t\|_2,\varepsilon)},
\quad \varepsilon=10^{-8}.
\label{eq:supp_target_embedding}
\end{equation}
Let $\mathcal{P}^{+}_g$, $\mathcal{P}^{-}_g$, and $\mathcal{P}^{0}$ denote the corresponding
sets of unit text embeddings.  CEAVAD computes the set-level support
\begin{equation}
E(\mathbf{q}_t,\mathcal{P})=
\log\left[\frac{1}{|\mathcal{P}|}
\sum_{\mathbf{p}\in\mathcal{P}}\exp(\mathbf{q}_t^{\top}\mathbf{p})\right]
\label{eq:supp_set_support}
\end{equation}
and the signed mechanism contrast
\begin{equation}
m_{t,g}=E(\mathbf{q}_t,\mathcal{P}^{+}_g)-
\max\{E(\mathbf{q}_t,\mathcal{P}^{0}),E(\mathbf{q}_t,\mathcal{P}^{-}_g)\}.
\label{eq:supp_margin}
\end{equation}
Thus the hazard bank must exceed both ordinary normality and its visually confusable benign
counterpart to obtain a positive raw margin.

Let $\gamma=\exp(\ell_{\mathrm{enc}})$ be the frozen encoder logit scale.  The composite
proposal
\begin{equation}
B_t=\log\left[\frac{1}{G}\sum_{g=1}^{G}
\exp(\gamma m_{t,g})\right]
\label{eq:supp_composite}
\end{equation}
aggregates the eight scaled mechanism margins without pruning the mechanism set.  CEAVAD uses
the native scale only for positively supported mechanisms when
the composite proposal is positive:
\begin{equation}
\widetilde m_{t,g}=
\begin{cases}
\gamma m_{t,g}, & B_t>0\ \text{and}\ m_{t,g}>0,\\
m_{t,g}, & \text{otherwise},
\end{cases}
\label{eq:supp_proposal_state}
\end{equation}
The bounded proposal activation is
$a_{t,g}=\sigma(\widetilde m_{t,g})$.
The resulting boundary proposal $\mathcal{R}_t$ contains the shared generic-normal account
and, for all eight mechanisms, $(H_g,N_g,\Phi_g,\widetilde m_{t,g},a_{t,g})$.  It also retains
$B_t$ and the raw margins when the composite proposal is positive.  All mechanisms remain
available to the subsequent visual adjudication regardless of rank.

\subsection{Partitioned Episodic Adjudication}

CEAVAD combines the proposal with an ordered episode
\begin{equation}
\mathcal{C}_t=(\widetilde X_t^{-},\widetilde X_t^{0},
\widetilde X_t^{+},\mathcal{R}_t),
\label{eq:supp_condition}
\end{equation}
where the three image groups are explicitly identified as \textsc{Past Context},
\textsc{Target Segment}, and \textsc{Future Context}.  The adjudication instruction marks the
target group as the sole scored interval and requires target-visible evidence to establish the
occurrence of a candidate hazard.  The adjacent groups then assess whether that target-visible
event violates local normality or is adequately accounted for by $N_g$ or $\mathcal{D}^{0}$.
Event-state descriptions connect visible onset, continuation, and resolution evidence across
these roles, while the instruction treats evidence confined to context as insufficient to
establish an anomaly in the target.

For leading-space, single-token continuations \texttt{abnormal} and \texttt{normal}, let
$z_{t,a}$ and $z_{t,n}$ be the candidate language-head logits.  The final interval posterior is
\begin{equation}
p_t=\frac{\exp(z_{t,a})}{\exp(z_{t,a})+\exp(z_{t,n})}.
\label{eq:supp_posterior}
\end{equation}
Each frame receives the score of its unique target interval, $s_f=p_t$ for $I_f\in W_t$.
The same condition $\mathcal{C}_t$ can be queried on demand for a compact account of the
target-visible evidence, temporal state, and resolution of the competing explanations.

\begin{algorithm}[t]
\caption{CEAVAD inference for one video.}
\label{alg:ceavad}
\begin{algorithmic}[1]
\REQUIRE Video $V$; fixed contrasts $\mathcal{M}$; frozen vision--language encoder and MLLM
\ENSURE Frame-aligned anomaly posteriors $\mathbf{s}(V)$
\STATE Divide $V$ into target intervals $\{W_t\}_{t=1}^{T}$ and sample target frames.
\STATE Embed every description bank once; retain the canonical pair and $\Phi_g$ for each $g$.
\FOR{$t=1$ to $T$}
    \STATE Compute $\mathbf{q}_t$ using Eq.~\ref{eq:supp_target_embedding}.
    \FOR{$g=1$ to $G$}
        \STATE Compute full-bank supports and $m_{t,g}$ using Eqs.~\ref{eq:supp_set_support}--\ref{eq:supp_margin}.
    \ENDFOR
    \STATE Compute $B_t$, $\widetilde m_{t,g}$, and $a_{t,g}$ using Eqs.~\ref{eq:supp_composite}--\ref{eq:supp_proposal_state}.
    \STATE Collect the adjacent \textsc{Past}/\textsc{Future} observations, assemble $\mathcal{R}_t$ with all mechanisms, and form $\mathcal{C}_t$.
    \STATE Evaluate the two continuation logits and obtain $p_t$ from Eq.~\ref{eq:supp_posterior}.
    \STATE Assign $p_t$ to every frame covered by $W_t$.
\ENDFOR
\RETURN $\mathbf{s}(V)$
\end{algorithmic}
\end{algorithm}

\section{Public-Safety Vocabulary and Prompting}
\label{sec:supp_vocabulary}

The fixed vocabulary is dataset-agnostic and is used unchanged for every target across all
benchmarks.  This
section reports its canonical competing statements, complete retrieval banks, event-state
specifications, and the condition presented to the MLLM.  The canonical hazard--benign pairs
are listed in Table~\ref{tab:supp_canonical_pairs}; the full retrieval banks are given in
Table~\ref{tab:supp_description_banks}.  The two
shared generic-normal descriptions are \emph{ordinary calm surveillance scene without danger}
and \emph{normal traffic walking standing shopping or routine activity}.  The event-state
specifications used to interpret ordered evidence appear in Table~\ref{tab:supp_event_states}.

\paragraph{Adjudication Condition.}

For each target, the visual message is ordered and role labeled as
\begin{quote}
\small
\textsc{Past Context} (context only; not the decision target): $\widetilde X_t^{-}$\\
\textsc{Target Segment} (the only decision target): $\widetilde X_t^{0}$\\
\textsc{Future Context} (context only; not the decision target): $\widetilde X_t^{+}$.
\end{quote}
The accompanying semantic condition contains the two generic-normal descriptions and one row
for every mechanism $g$.  Each row has the fields
\begin{quote}
\small
\texttt{hazard mechanism}: $g$; \texttt{hazard hypothesis}: $H_g$;\\
\texttt{benign counterpart}: $N_g$; \texttt{event state}: $\Phi_g$;\\
\texttt{proposal margin}: $\widetilde m_{t,g}$; \texttt{activation}: $a_{t,g}$.
\end{quote}
When $B_t>0$, the condition additionally exposes $B_t$ and the eight raw margins
$\{m_{t,g}\}_{g=1}^{G}$ so that the MLLM can distinguish the composite proposal from its
mechanism-level support.  The eight rows are retained in their fixed order; no row is removed
according to its activation or rank.

The adjudication instruction assigns occurrence and local-normality evidence to different
temporal roles: visible people, objects, actions, relations, or consequences in the
\textsc{Target} establish whether an event occurs in the scored interval; the adjacent groups
assess whether that target-visible event violates the local episode or remains compatible with
an ordinary or mechanism-specific benign account.  The hazard explanation is accepted only
when a concrete hazard occurs in the target and the available benign account is inadequate.
The assistant state is prefixed by
\begin{quote}
\small\raggedright\ttfamily\sloppy
After comparing all competing semantic-memory explanations and benign alternatives only against
TARGET-visible evidence, the TARGET segment is
\end{quote}
Here, ``TARGET-visible evidence'' constrains event occurrence: adjacent observations remain
available to assess local normality and benign adequacy, but cannot supply an event absent from
the target.
The continuation is constrained to the leading-space, single-token candidates
\texttt{ abnormal} and \texttt{ normal}.  Their candidate logits define the posterior in
Eq.~\ref{eq:supp_posterior}.

\section{Implementation and Evaluation Details}
\label{sec:supp_implementation}

For benchmark evaluation, the interval posterior from Eq.~\ref{eq:supp_posterior} is assigned
to every video frame covered by its target interval.  This produces the frame-aligned sequence
used by the benchmark metrics and by the qualitative cases in
Fig.~\ref{fig:supp_qualitative_cross_dataset}.

For a video with frame rate $r$ and target interval $[u,v)$, the corresponding frame range is
$[\lfloor ru\rfloor,\lceil rv\rceil)$ after clipping to the valid video extent.  The
nonoverlapping intervals therefore give each frame one score; the final interval is truncated
at the video end.  The evaluated records use 30 fps on UCF-Crime and UBnormal and 24 fps on
XD-Violence, so the two-second boundaries map to disjoint integer frame ranges.

For UCF-Crime, the official abnormal frame ranges are rasterized into a
binary frame sequence and ROC-AUC is computed after concatenating all 290 test videos.
UBnormal uses the provided test annotations and reports micro frame-level ROC-AUC over all
211 videos.  For XD-Violence, official
temporal ranges are rasterized for all 800 test videos; we report concatenated frame-level
ROC-AUC and average precision (AP), with AP as the primary metric.  The same interval
posteriors and frame mapping are used for every method variant.

Beyond the two-token continuation in Eq.~\ref{eq:supp_posterior}, which provides the anomaly
score, CEAVAD reuses the same role-separated episode and boundary proposal to obtain a
structured evidence-grounded account of the target event.  The account identifies the
target-visible event and evidence, its temporal state, and the benign explanation considered
by the adjudication, as illustrated in Fig.~\ref{fig:supp_qualitative_cross_dataset}.

\setcounter{table}{3}
\section{Extended Ablation Studies}
\label{sec:supp_ablations}

We extend the main-paper ablation by separating the scientific roles of event-contrast
construction, proposal formation, mechanism retention, and temporally partitioned evidence.
The controlled variants and their results are given in Table~\ref{tab:supp_ablation_two_dataset}.

Table~\ref{tab:supp_ablation_two_dataset} extends the main-paper component study with controlled proposal-formation and temporal-evidence variants.  All conditions use the same target geometry, frozen models, test sets, and two-token scoring rule; all CEAVAD variants retain the fixed mechanism vocabulary, whereas Direct Qwen omits the event contrasts and semantic proposal.
\emph{Direct Qwen} removes the event contrasts and semantic proposal.  For proposal formation, \emph{Unscaled margins} supplies the raw $m_{t,g}$ values and their sigmoid activations without native scaling or composite gating, whereas \emph{Unconditional scaling} applies $\gamma$ to every positive margin without the $B_t>0$ condition.  For temporal evidence, \emph{Target only} uses the complete proposal with target frames alone; \emph{Unpartitioned episode} presents the target and ordered context as one visual sequence; and \emph{w/o evidence partition} labels the three temporal roles but lets evidence from all three contribute without assigning distinct evidential responsibilities.  \emph{Top-1 mechanism} retains only the highest raw-margin mechanism, while \emph{w/o event-state} removes $\Phi_g$ from the otherwise complete partitioned adjudication.

\begin{table}[H]
\centering
\small
\begin{tabular*}{\columnwidth}{@{\extracolsep{\fill}}lcc@{}}
\toprule
Condition & UCF-Crime & UBnormal \\
& ROC-AUC (\%) & ROC-AUC (\%) \\
\midrule
Direct Qwen & 80.60 & 70.41 \\
Unscaled margins & 80.88 & 74.20 \\
Unconditional scaling & 81.57 & 72.94 \\
Target only & 84.32 & 73.82 \\
Unpartitioned episode & 83.99 & 73.65 \\
w/o evidence partition & 85.11 & 75.58 \\
Top-1 mechanism & 83.42 & 71.43 \\
w/o event-state & 84.28 & 74.01 \\
\textbf{Full CEAVAD} & \textbf{86.15} & \textbf{77.55} \\
\bottomrule
\end{tabular*}
\caption{Key ablations on UCF-Crime and UBnormal.}
\label{tab:supp_ablation_two_dataset}
\end{table}

Full CEAVAD exceeds Direct Qwen by 5.55 ROC-AUC points on UCF-Crime and 7.14 points on UBnormal.  The component pattern is consistent but more pronounced on UBnormal: retaining only the top mechanism reduces ROC-AUC by 2.73 points on UCF-Crime and 6.12 points on UBnormal, while removing event-state descriptions costs 1.87 and 3.54 points, respectively.  The larger UBnormal drops indicate greater performance dependence on alternative mechanisms for visual revision and event-evolution modeling.  The temporal controls isolate a separate effect: target-only, unpartitioned, and non-role-specific episodes lose 1.04--2.16 points on UCF-Crime and 1.97--3.90 points on UBnormal, showing that both surrounding context and its evidential partition contribute.  The proposal-formation controls show the same pattern: unscaled margins trail Full CEAVAD by 5.27 and 3.35 points, and unconditional scaling by 4.58 and 4.61 points, on UCF-Crime and UBnormal, respectively.  Together, these controls support complementary roles for proposal construction, mechanism retention, event-state evidence, and partitioned context.

\section{Robustness and Sensitivity}
\label{sec:supp_robustness}

This section examines temporal geometry and textual realization.
Each controlled study retains the complete CEAVAD adjudication procedure and varies only
the stated factor.

\subsection{Temporal Window Robustness}

\begin{figure}[H]
\centering
\includegraphics[width=\columnwidth]{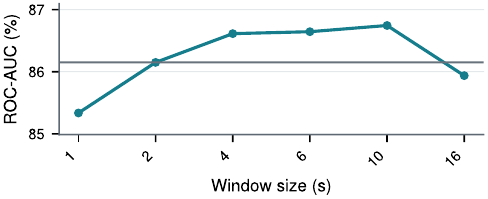}
\caption{Sensitivity to temporal window size on UCF-Crime.}
\label{fig:supp_temporal}
\end{figure}

We vary one temporal parameter, the window size, while retaining the fixed event vocabulary, frozen models, and adjudication rule.  Changing the window size changes the nonoverlapping target interval and its stride together.  Sampling remains fixed at 1 fps, so the number of observations in the target and context groups changes deterministically with the window size.  Table~\ref{tab:supp_temporal} and Figure~\ref{fig:supp_temporal} report the complete-test-set results.

\begin{table}[H]
\centering
\small
\begin{tabular*}{\columnwidth}{@{\extracolsep{\fill}}cc@{}}
\toprule
Window size (s) & ROC-AUC (\%) \\
\midrule
1 & 85.34 \\
2 (default) & \textbf{86.15} \\
4 & 86.61 \\
6 & 86.64 \\
10 & 86.74 \\
16 & 85.94 \\
\bottomrule
\end{tabular*}
\caption{Sensitivity to temporal window size on UCF-Crime.}
\label{tab:supp_temporal}
\end{table}

Across window sizes from one to sixteen seconds, UCF-Crime ROC-AUC remains within 85.34--86.74.  The two-second default reaches 86.15, exactly 0.59 points below the displayed maximum of 86.74, and the alternatives show no monotonic dependence on window size.

\subsection{Robustness to Textual Realization}

\begin{figure*}[t]
\centering
\includegraphics[width=0.94\textwidth]{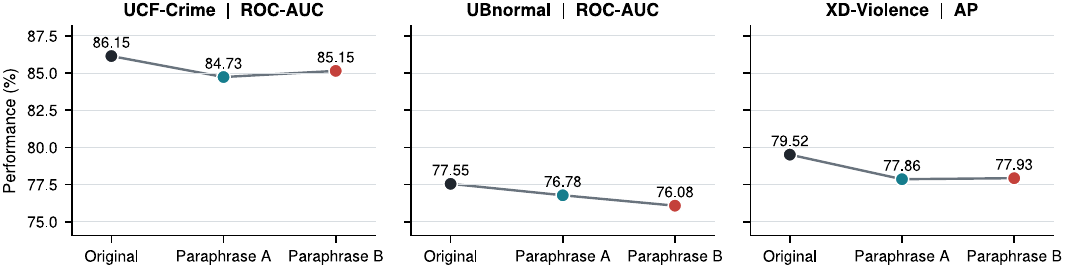}
\caption{Primary metrics across frozen textual realizations.}
\label{fig:supp_paraphrases}
\end{figure*}

We test whether CEAVAD depends on one exact wording of its fixed event vocabulary.  Before
evaluation, we froze two dataset-agnostic alternative realizations that retain the same eight
mechanisms, hazard--benign polarity, three descriptions per side, two generic-normal
descriptions, and unchanged event-state specifications.  They change only the wording of the
retrieval descriptions and canonical competing statements.  Table~\ref{tab:supp_paraphrases}
and Fig.~\ref{fig:supp_paraphrases} report complete test-set results for all three banks.  The
two alternative banks were constructed as dataset-agnostic paraphrases before evaluation: each
keeps the same eight mechanism identities, three hazard descriptions, three mechanism-specific
benign descriptions, and event-state specifications, while its two generic-normal descriptions
are frozen with that realization.  Once fixed, each bank is used unchanged for every target and
benchmark.  Their complete contents are reproduced in the unnumbered end tables for
\hyperlink{supp-bank-a}{Paraphrase A} and \hyperlink{supp-bank-b}{Paraphrase B}.

\begin{table}[H]
\centering
\small
\begin{tabular*}{\columnwidth}{@{\extracolsep{\fill}}lcccc@{}}
\toprule
& UCF & UB & \multicolumn{2}{c}{XD} \\
\cmidrule(lr){4-5}
Text bank & \shortstack{ROC-AUC\\(\%)} & \shortstack{ROC-AUC\\(\%)} & \shortstack{ROC-AUC\\(\%)} & \shortstack{AP\\(\%)} \\
\midrule
Original & 86.15 & 77.55 & 93.40 & 79.52 \\
Paraphrase A & 84.73 & 76.78 & 93.01 & 77.86 \\
Paraphrase B & 85.15 & 76.08 & 92.75 & 77.93 \\
\bottomrule
\end{tabular*}
\caption{Performance across frozen textual realizations.}
\label{tab:supp_paraphrases}
\end{table}

Across the original and two paraphrased realizations, the primary-metric ranges are
1.42, 1.47, and 1.66 points on UCF-Crime,
UBnormal, and XD-Violence, respectively.  The bounded variation across all three benchmarks
indicates limited sensitivity to the exact surface wording under these frozen realizations,
while mechanism membership, polarity, and event-state semantics remain fixed.

\section{Additional Qualitative Analysis}
\label{sec:supp_qualitative}

Target-centered cases cover confirmed hazards, visually confusable benign events,
evidence-recovered hazards, and temporal transitions.

\begin{figure*}[!t]
\centering
\includegraphics[width=0.94\textwidth]{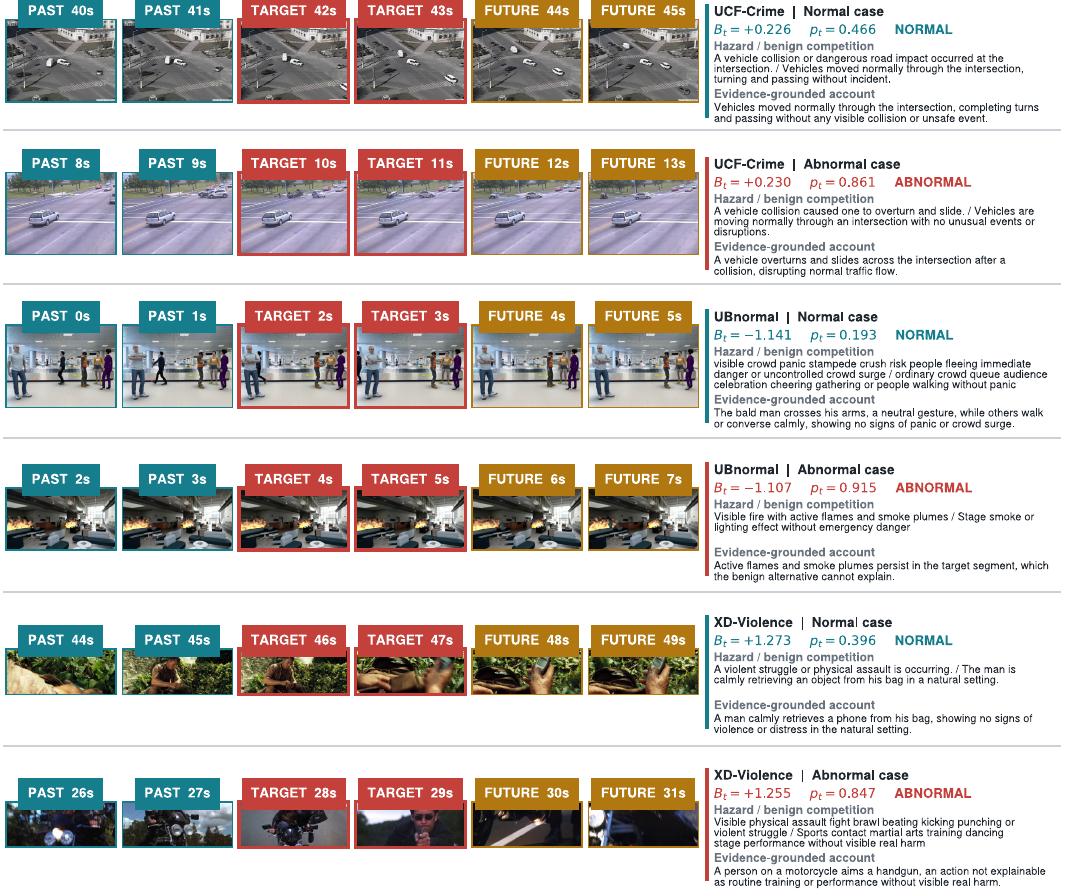}
\caption{Evidence-grounded event adjudication across three VAD benchmarks.}
\label{fig:supp_qualitative_cross_dataset}
\par\smallskip
\begin{minipage}{0.94\textwidth}
\small
\textit{Case reading.}  The rows pair ordinary and hazardous targets from UCF-Crime,
UBnormal, and XD-Violence.  The target-centered frames, proposal, posterior, and tested
competition are shown together so that the reported decision can be traced to visible
evidence.  Controlled vehicle motion and routine crowd or object activity support the
ordinary cases, whereas target-visible collision, fire, and weapon threat support the
corresponding hazardous cases.  On UCF-Crime, the ordinary vehicle sequence remains
orderly while the paired target contains an overturning vehicle.  On UBnormal, the
ordinary crowd remains calm while the paired target contains persistent flames and smoke.
On XD-Violence, routine object handling contrasts with a target containing an aimed
handgun.  These examples expose the event evidence and the competing account in the same
visual record.
\end{minipage}
\end{figure*}

Figure~\ref{fig:supp_qualitative_cross_dataset} pairs one normal and one abnormal target from
each benchmark, none of which appears in the main-paper case study.  Each row shows the
two-frame \textsc{Past}, \textsc{Target}, and \textsc{Future} groups, proposal $B_t$, posterior
$p_t$, tested competition, and an account generated from the same condition used for scoring;
every statement is restricted to evidence visible in the displayed frames.

On UCF-Crime, controlled vehicle motion sustains the benign account, while target-visible
impact and an overturning vehicle confirm the competing hazard.  On UBnormal, ordinary
movement provides no collective-distress evidence, whereas persistent flame and smoke confirm
an active fire.  On XD-Violence, routine object handling supports the benign account, while an
aimed handgun supplies target-visible evidence that the competing benign explanation cannot
accommodate.

The XD-Violence benign case has $B_t=1.273$ but $p_t=0.396$, whereas the UBnormal fire
case has $B_t=-1.107$ and $p_t=0.915$.  These opposite revisions show that video evidence
can reject a strong semantic proposal or confirm a hazard with weak initial retrieval support.

\section{Limitations}
\label{sec:supp_limitations}

CEAVAD addresses video anomaly detection and understanding with a fixed eight-mechanism
hypothesis space that is used unchanged across all three benchmarks, but no compact vocabulary
can exhaust the open set of possible anomalies; specialized environments may require
additional hazard--benign contrasts.  The current implementation also uses compact,
target-centered episodes sampled at one frame per second.  This design captures local event
evolution, while anomalies defined by very brief evidence, long-range dependencies, or
nonvisual signals may benefit from adaptive sampling, longer temporal context, or additional
modalities such as audio.  Finally, although CEAVAD requires no target-domain training,
its interval-wise adjudication leaves room for greater inference efficiency on long videos.
Reusing shared evidence across neighboring intervals and improving inference efficiency are
promising directions for future work.

\clearpage
\setcounter{table}{0}
\begin{table*}[t]
\centering
\small
\begin{tabular}{@{}>{\raggedright\arraybackslash}p{0.25\textwidth}p{0.345\textwidth}p{0.345\textwidth}@{}}
\toprule
Mechanism & Canonical hazard statement $H_g$ & Canonical benign counterpart $N_g$ \\
\midrule
Physical violence and weapon threat & visible physical assault fight brawl beating kicking punching or violent struggle & sports contact martial arts training dancing stage performance without visible real harm \\
Vehicle and mobility hazards & visible vehicle crash collision rollover person hit by vehicle or dangerous road impact & ordinary traffic congestion braking turning parking waiting at intersection or vehicles moving normally \\
Fire, explosion, and hazardous release & visible fire flame heavy smoke explosion blast burning object or hazardous smoke plume & fog steam dust stage smoke lighting effect fireworks or visual effects without emergency danger \\
Intrusion and property violation & visible burglary forced entry theft robbery shoplifting vandalism or property damage & normal shopping delivery carrying bags maintenance cleaning moving boxes or opening doors normally \\
Crowd panic and collective disorder & visible crowd panic stampede crush risk people fleeing immediate danger or uncontrolled crowd surge & ordinary crowd queue audience celebration cheering gathering or people walking without panic \\
Human distress and medical emergency & visible person falling collapsing lying injured unconscious having seizure or unable to stand & person sitting resting exercising stretching sleeping in normal context or bending to pick something up \\
Restricted-area and safety-rule violation & visible trespassing restricted area entry unsafe climbing wrong-way movement or dangerous rule violation & authorized worker guard resident staff maintenance or emergency responder in restricted-looking area \\
Environmental, structural, and industrial hazards & visible structural collapse falling object electrocution machinery entrapment or industrial safety hazard & normal construction maintenance machinery operation worker activity or equipment movement without danger \\
\bottomrule
\end{tabular}
\caption{Canonical hazard--benign event competitions used by the MLLM.}
\label{tab:supp_canonical_pairs}
\end{table*}

\begin{table*}[p]
\centering
\footnotesize
\setlength{\tabcolsep}{3pt}
\renewcommand{\arraystretch}{0.96}
\begin{tabular}{@{}>{\raggedright\arraybackslash}p{0.155\textwidth}p{0.395\textwidth}p{0.395\textwidth}@{}}
\toprule
Mechanism & Complete hazard bank $\mathcal{D}^{+}_g$ & Complete mechanism-specific benign bank $\mathcal{D}^{-}_g$ \\
\midrule
Physical violence and weapon threat & \begin{enumerate}\setlength{\itemsep}{0pt}\setlength{\parskip}{0pt}\setlength{\parsep}{0pt}\setlength{\topsep}{0pt}\item visible physical assault fight brawl beating kicking punching or violent struggle\item visible weapon threat shooting stabbing armed robbery or person threatening another person\item visible person injured by deliberate attack or forced violent contact\end{enumerate} & \begin{enumerate}\setlength{\itemsep}{0pt}\setlength{\parskip}{0pt}\setlength{\parsep}{0pt}\setlength{\topsep}{0pt}\item sports contact martial arts training dancing stage performance without visible real harm\item people arguing gesturing playing acting or rehearsing without physical injury or weapon threat\item police responders or crowd watching without active assault happening now\end{enumerate} \\
Vehicle and mobility hazards & \begin{enumerate}\setlength{\itemsep}{0pt}\setlength{\parskip}{0pt}\setlength{\parsep}{0pt}\setlength{\topsep}{0pt}\item visible vehicle crash collision rollover person hit by vehicle or dangerous road impact\item visible damaged vehicle debris blocked road traffic accident injury fire or smoke after collision\item visible unsafe mobility event such as illegal turn near collision jaywalking danger or vehicle out of lane\end{enumerate} & \begin{enumerate}\setlength{\itemsep}{0pt}\setlength{\parskip}{0pt}\setlength{\parsep}{0pt}\setlength{\topsep}{0pt}\item ordinary traffic congestion braking turning parking waiting at intersection or vehicles moving normally\item headlights reflections dust rain shadows road motion or stopped cars without collision or injury\item pedestrians crossing normally cyclists riding in lane or vehicles maneuvering safely\end{enumerate} \\
Fire, explosion, and hazardous release & \begin{enumerate}\setlength{\itemsep}{0pt}\setlength{\parskip}{0pt}\setlength{\parsep}{0pt}\setlength{\topsep}{0pt}\item visible fire flame heavy smoke explosion blast burning object or hazardous smoke plume\item visible dangerous fire explosion aftermath debris emergency damage or active burning hazard\item visible hazardous material leak chemical smoke gas cloud or unsafe environmental condition\end{enumerate} & \begin{enumerate}\setlength{\itemsep}{0pt}\setlength{\parskip}{0pt}\setlength{\parsep}{0pt}\setlength{\topsep}{0pt}\item fog steam dust stage smoke lighting effect fireworks or visual effects without emergency danger\item normal cooking smoke exhaust mist haze reflection or weather-related low visibility\item firefighters responders equipment lights or smoke-like background without visible active hazard\end{enumerate} \\
Intrusion and property violation & \begin{enumerate}\setlength{\itemsep}{0pt}\setlength{\parskip}{0pt}\setlength{\parsep}{0pt}\setlength{\topsep}{0pt}\item visible burglary forced entry theft robbery shoplifting vandalism or property damage\item visible person stealing snatching breaking door breaking window damaging property or threatening staff\item visible unauthorized intrusion into restricted private or closed area with suspicious handling of objects\end{enumerate} & \begin{enumerate}\setlength{\itemsep}{0pt}\setlength{\parskip}{0pt}\setlength{\parsep}{0pt}\setlength{\topsep}{0pt}\item normal shopping delivery carrying bags maintenance cleaning moving boxes or opening doors normally\item authorized worker staff customer resident entering leaving browsing shelves or handling goods routinely\item people exchanging objects talking waiting repairing or transporting property without theft or damage\end{enumerate} \\
Crowd panic and collective disorder & \begin{enumerate}\setlength{\itemsep}{0pt}\setlength{\parskip}{0pt}\setlength{\parsep}{0pt}\setlength{\topsep}{0pt}\item visible crowd panic stampede crush risk people fleeing immediate danger or uncontrolled crowd surge\item visible riot violent crowd disorder people running in fear pushing falling or being crushed\item visible emergency crowd evacuation chaos with injured people pressure or blocked escape\end{enumerate} & \begin{enumerate}\setlength{\itemsep}{0pt}\setlength{\parskip}{0pt}\setlength{\parsep}{0pt}\setlength{\topsep}{0pt}\item ordinary crowd queue audience celebration cheering gathering or people walking without panic\item sports fans dancers performers parade festival or busy street crowd without visible danger\item people running for exercise commuting hurry or play without fear injury pushing or collapse\end{enumerate} \\
Human distress and medical emergency & \begin{enumerate}\setlength{\itemsep}{0pt}\setlength{\parskip}{0pt}\setlength{\parsep}{0pt}\setlength{\topsep}{0pt}\item visible person falling collapsing lying injured unconscious having seizure or unable to stand\item visible medical emergency human distress person trapped struck crushed or needing urgent help\item visible body on ground after hazard injury impact or abnormal immobility in unsafe context\end{enumerate} & \begin{enumerate}\setlength{\itemsep}{0pt}\setlength{\parskip}{0pt}\setlength{\parsep}{0pt}\setlength{\topsep}{0pt}\item person sitting resting exercising stretching sleeping in normal context or bending to pick something up\item actor performer athlete dancer or child playing on ground without visible distress or injury\item people waiting lying on beach bench floor mat or staged scene without emergency signs\end{enumerate} \\
Restricted-area and safety-rule violation & \begin{enumerate}\setlength{\itemsep}{0pt}\setlength{\parskip}{0pt}\setlength{\parsep}{0pt}\setlength{\topsep}{0pt}\item visible trespassing restricted area entry unsafe climbing wrong-way movement or dangerous rule violation\item visible person or vehicle in forbidden unsafe zone such as tracks road lane roof construction area or closed property\item visible suspicious loitering near restricted entrance security boundary or unsafe facility context\end{enumerate} & \begin{enumerate}\setlength{\itemsep}{0pt}\setlength{\parskip}{0pt}\setlength{\parsep}{0pt}\setlength{\topsep}{0pt}\item authorized worker guard resident staff maintenance or emergency responder in restricted-looking area\item normal pedestrian crossing at crosswalk vehicle turning legally or person waiting in allowed area\item tourist visitor delivery worker or routine inspection without visible danger or unauthorized behavior\end{enumerate} \\
Environmental, structural, and industrial hazards & \begin{enumerate}\setlength{\itemsep}{0pt}\setlength{\parskip}{0pt}\setlength{\parsep}{0pt}\setlength{\topsep}{0pt}\item visible structural collapse falling object electrocution machinery entrapment or industrial safety hazard\item visible severe weather flood debris falling sign broken structure or environmental danger threatening people\item visible person caught trapped crushed struck by object or endangered by machinery equipment or collapse\end{enumerate} & \begin{enumerate}\setlength{\itemsep}{0pt}\setlength{\parskip}{0pt}\setlength{\parsep}{0pt}\setlength{\topsep}{0pt}\item normal construction maintenance machinery operation worker activity or equipment movement without danger\item ordinary rain wind shadows moving flags water reflection or background clutter without structural hazard\item stored materials tools vehicles machines or buildings appearing intact without visible accident\end{enumerate} \\
\bottomrule
\end{tabular}
\caption{Complete description banks used for visual--language retrieval.}
\label{tab:supp_description_banks}
\end{table*}

\begin{table*}[p]
\centering
\small
\setlength{\tabcolsep}{3pt}
\renewcommand{\arraystretch}{0.96}
\begin{tabular}{@{}>{\raggedright\arraybackslash}p{0.19\textwidth}p{0.765\textwidth}@{}}
\toprule
Mechanism & Event-state specification $\Phi_g$ \\
\midrule
Physical violence and weapon threat & \textbf{Unsafe state:} a person remains under active violent force, coercion, or weapon threat\newline \textbf{Onset:} a strike, forced contact, pursuit, restraint, or weapon threat begins\newline \textbf{Continuation:} the attack, struggle, coercion, injury, or weapon control remains visible\newline \textbf{Termination or benign resolution:} contact has ended and visible interaction is sport, performance, assistance, or routine social activity \\
Vehicle and mobility hazards & \textbf{Unsafe state:} a collision, dangerous vehicle-person relation, or uncontrolled mobility hazard remains active\newline \textbf{Onset:} impact, loss of control, unsafe incursion, or a near-collision relation begins\newline \textbf{Continuation:} impact motion, a person in the vehicle path, wreckage, blockage, or dangerous trajectory remains visible\newline \textbf{Termination or benign resolution:} traffic is controlled and visible motion is ordinary braking, parking, turning, crossing, or lane-following \\
Fire, explosion, and hazardous release & \textbf{Unsafe state:} active fire, hazardous smoke, explosion damage, or dangerous material release remains present\newline \textbf{Onset:} ignition, blast, rupture, sparking, or hazardous release begins\newline \textbf{Continuation:} flame, dense sourced smoke, spreading plume, debris, damage, or exposure remains visible\newline \textbf{Termination or benign resolution:} the source is absent or resolved and the appearance is fog, steam, dust, exhaust, weather, or a visual effect \\
Intrusion and property violation & \textbf{Unsafe state:} unauthorized access, dispossession, concealment, forced entry, or property damage remains in progress\newline \textbf{Onset:} forced access, snatching, concealment, removal, breaking, or unauthorized boundary crossing begins\newline \textbf{Continuation:} the taking, intrusion, escape with property, forced access, or damage remains visibly connected\newline \textbf{Termination or benign resolution:} visible activity is authorized shopping, delivery, maintenance, repair, access, or ordinary object handling \\
Crowd panic and collective disorder & \textbf{Unsafe state:} uncontrolled collective flight, violent disorder, crush pressure, or evacuation danger remains active\newline \textbf{Onset:} fear-driven flight, violent crowd surge, pushing, falling, or emergency evacuation begins\newline \textbf{Continuation:} panic directionality, compression, uncontrolled surge, injury, or blocked escape remains visible\newline \textbf{Termination or benign resolution:} the group is orderly and visible movement is a queue, celebration, performance, commute, or exercise \\
Human distress and medical emergency & \textbf{Unsafe state:} a person remains fallen, trapped, injured, unresponsive, or in visible medical distress\newline \textbf{Onset:} a fall, collapse, impact, seizure, entrapment, or loss of mobility begins\newline \textbf{Continuation:} abnormal immobility, inability to rise, injury, entrapment, or urgent assistance remains visible\newline \textbf{Termination or benign resolution:} the person visibly recovers or the posture is resting, exercise, play, performance, or another ordinary activity \\
Restricted-area and safety-rule violation & \textbf{Unsafe state:} unsafe or unauthorized occupation of a hazardous restricted path or area remains active\newline \textbf{Onset:} a person or vehicle enters, climbs into, or moves against a visibly restricted or dangerous path\newline \textbf{Continuation:} continued unsafe occupancy, climbing, wrong-way motion, boundary crossing, or exposure remains visible\newline \textbf{Termination or benign resolution:} the subject is visibly authorized or follows an allowed crossing, work, inspection, delivery, or routine access path \\
Environmental, structural, and industrial hazards & \textbf{Unsafe state:} structural, mechanical, electrical, falling-object, or severe environmental danger remains active\newline \textbf{Onset:} collapse, entrapment, machinery contact, falling debris, electrocution, flooding, or severe exposure begins\newline \textbf{Continuation:} instability, entrapment, dangerous machinery relation, debris, damage, flooding, or threatened people remains visible\newline \textbf{Termination or benign resolution:} structures and equipment are intact and activity is controlled construction, maintenance, operation, or ordinary weather \\
\bottomrule
\end{tabular}
\caption{Complete event-state descriptions used for evidence adjudication.}
\label{tab:supp_event_states}
\end{table*}

\newcommand{\suppbanklist}[3]{%
  \begin{minipage}[t]{\linewidth}%
  \begin{enumerate}[label=\arabic*.,leftmargin=1.05em,itemsep=0pt,topsep=0pt,parsep=0pt]%
  \item #1
  \item #2
  \item #3
  \end{enumerate}%
  \end{minipage}%
}

\begin{table*}[p]
\centering
\footnotesize
\setlength{\tabcolsep}{3pt}
\renewcommand{\arraystretch}{0.94}
\hypertarget{supp-bank-a}{}
\begin{tabular}{@{}>{\raggedright\arraybackslash}p{0.155\textwidth}p{0.395\textwidth}p{0.395\textwidth}@{}}
\toprule
Mechanism & Paraphrase A hazard bank $\mathcal{D}^{+}_{g,A}$ & Paraphrase A benign bank $\mathcal{D}^{-}_{g,A}$ \\
\midrule
\multicolumn{3}{@{}p{0.97\textwidth}@{}}{\textbf{Generic-normal descriptions:} a calm ordinary scene with no visible public-safety threat; routine walking standing traffic shopping or other normal activity.} \\
\midrule
Physical violence and weapon threat &
\suppbanklist{a visible physical attack including fighting beating kicking punching or a violent struggle}{a visible firearm or blade threat shooting stabbing armed robbery or direct threat against a person}{a visible injury caused by an intentional attack or forceful violent contact} &
\suppbanklist{visible sports contact martial-arts practice dance or stage action without real harm}{visible arguing gesturing play acting or rehearsal without injury or weapon threat}{police emergency personnel or onlookers present while no assault is visibly occurring} \\
Vehicle and mobility hazards &
\suppbanklist{a visible road collision rollover pedestrian impact or other dangerous vehicle crash}{visible crash aftermath with a damaged vehicle debris blocked road injury fire or smoke}{visible unsafe mobility such as wrong-lane travel dangerous crossing or an imminent collision} &
\suppbanklist{normal traffic slowing turning parking waiting or moving without a crash}{headlights rain dust shadows road motion or stopped vehicles with no impact or injury}{pedestrians cross normally cyclists remain in lane or vehicles maneuver under control} \\
Fire, explosion, and hazardous release &
\suppbanklist{visible flames active burning dense smoke an explosion or a hazardous plume}{visible blast or fire aftermath with debris damage smoke or an active ignition source}{a visible chemical or gas leak unsafe vapor cloud or other hazardous material release} &
\suppbanklist{visible fog steam dust stage smoke fireworks lighting or visual effects without danger}{cooking smoke exhaust mist haze reflections or weather-related low visibility}{responders equipment warning lights or smoke-like background without an active hazard} \\
Intrusion and property violation &
\suppbanklist{visible theft robbery burglary forced entry vandalism or property destruction}{a person visibly steals snatches breaks access damages property or threatens staff}{unauthorized entry into private or restricted space with suspicious object handling} &
\suppbanklist{routine shopping delivery bag carrying cleaning maintenance or normal use of a doorway}{authorized staff customers residents or workers enter leave browse or handle goods normally}{people exchange carry repair or transport objects without theft damage or coercion} \\
Crowd panic and collective disorder &
\suppbanklist{visible crowd panic stampede crushing pressure or people fleeing an immediate threat}{visible riot or uncontrolled crowd movement with fear pushing falls or crush danger}{a visible emergency evacuation with chaos injury crowd pressure or blocked escape} &
\suppbanklist{an orderly queue audience celebration gathering or walking crowd without panic}{sports spectators dancers performers parade or festival crowds without visible danger}{people run for exercise commuting urgency or play without fear pushing injury or collapse} \\
Human distress and medical emergency &
\suppbanklist{a visible person falls collapses lies injured becomes unconscious has a seizure or cannot stand}{a visible medical emergency with a trapped struck crushed or severely distressed person needing help}{a body remains on the ground after impact injury or another hazard in an unsafe context} &
\suppbanklist{a person sits rests stretches exercises sleeps normally or bends to retrieve an object}{an actor athlete dancer performer or child is on the ground without distress or injury}{people lie or wait normally on a beach bench floor or mat without emergency evidence} \\
Restricted-area and safety-rule violation &
\suppbanklist{visible trespass entry into a restricted area unsafe climbing wrong-way travel or dangerous rule breaking}{a person or vehicle visibly occupies a forbidden hazardous zone such as tracks a road lane roof or closed site}{visible suspicious lingering or boundary crossing near a restricted entrance or unsafe facility} &
\suppbanklist{authorized staff guards residents maintenance workers or responders occupy a restricted-looking area}{pedestrians use a crosswalk vehicles turn legally or people wait in an allowed location}{visitors delivery workers tourists or inspectors follow an ordinary authorized access path} \\
Environmental, structural, and industrial hazards &
\suppbanklist{visible structural failure falling debris electrical danger machinery entrapment or an industrial accident}{visible flood severe weather broken structure or falling object that threatens people}{a person is visibly trapped crushed struck or endangered by machinery equipment or collapse} &
\suppbanklist{controlled construction maintenance machinery use or worker activity without visible danger}{ordinary rain wind shadows moving flags water reflections or clutter without structural failure}{stored tools materials vehicles machines or intact buildings with no visible accident} \\
\bottomrule
\end{tabular}
\caption*{\textbf{Paraphrase A bank.} Complete frozen hazard, mechanism-specific benign, and generic-normal descriptions used in the textual-realization control.}
\end{table*}

\begin{table*}[p]
\centering
\footnotesize
\setlength{\tabcolsep}{3pt}
\renewcommand{\arraystretch}{0.94}
\hypertarget{supp-bank-b}{}
\begin{tabular}{@{}>{\raggedright\arraybackslash}p{0.155\textwidth}p{0.395\textwidth}p{0.395\textwidth}@{}}
\toprule
Mechanism & Paraphrase B hazard bank $\mathcal{D}^{+}_{g,B}$ & Paraphrase B benign bank $\mathcal{D}^{-}_{g,B}$ \\
\midrule
\multicolumn{3}{@{}p{0.97\textwidth}@{}}{\textbf{Generic-normal descriptions:} people and vehicles follow ordinary controlled activity without visible danger; routine movement waiting work shopping or social interaction continues normally.} \\
\midrule
Physical violence and weapon threat &
\suppbanklist{one person visibly strikes kicks punches restrains or violently grapples with another}{a weapon is visibly used or aimed while a person is shot stabbed robbed or threatened}{deliberate force visibly injures or overpowers a person} &
\suppbanklist{participants make controlled contact in sport training dance or a staged performance}{people argue gesture play or rehearse without injury coercion or a weapon threat}{responders or bystanders are present but no active attack is visible} \\
Vehicle and mobility hazards &
\suppbanklist{a vehicle visibly collides rolls over or strikes a person or object}{impact leaves visible damage debris blockage injury fire or smoke around a vehicle}{a vehicle or pedestrian enters an unsafe path with imminent collision risk} &
\suppbanklist{vehicles brake turn park queue or travel normally without impact}{light weather dust shadow or camera motion resembles an incident but no collision is visible}{pedestrians cyclists and vehicles move through their expected paths under control} \\
Fire, explosion, and hazardous release &
\suppbanklist{visible flame active burning dense sourced smoke or an explosion is present}{a blast or fire leaves visible debris damage smoke or continuing combustion}{gas vapor chemical material or another hazardous substance visibly escapes} &
\suppbanklist{fog steam dust stage smoke fireworks or lighting creates a harmless smoke-like appearance}{cooking exhaust mist haze reflection or weather explains the visible plume}{responders or equipment are visible but no active fire blast or release is present} \\
Intrusion and property violation &
\suppbanklist{a person visibly forces entry steals robs vandalizes or damages property}{a person snatches conceals removes or breaks property without authorization}{a person crosses a protected boundary and handles objects in an unauthorized manner} &
\suppbanklist{people shop deliver clean maintain carry goods or use doors routinely}{authorized staff customers residents or workers handle property normally}{objects are exchanged repaired moved or transported without forced access or theft} \\
Crowd panic and collective disorder &
\suppbanklist{a crowd visibly flees surges or compresses under panic or immediate danger}{violent disorder causes uncontrolled pushing falling or crush pressure}{an emergency evacuation visibly produces chaos injury or blocked escape} &
\suppbanklist{the crowd queues gathers cheers or walks in an orderly way}{spectators dancers performers or festival participants move without danger}{people run for travel exercise or play without panic pressure or collapse} \\
Human distress and medical emergency &
\suppbanklist{a person visibly falls collapses convulses lies injured or cannot stand}{a trapped struck crushed or unresponsive person visibly requires urgent help}{a person remains abnormally motionless on the ground after impact or danger} &
\suppbanklist{a person sits rests sleeps stretches exercises or bends normally}{a performer athlete dancer or child lies or plays on the ground without distress}{a person waits or reclines in an ordinary setting without injury evidence} \\
Restricted-area and safety-rule violation &
\suppbanklist{a person or vehicle visibly enters climbs into or moves against a restricted unsafe path}{a subject remains in a forbidden zone such as tracks traffic lanes roofs or closed property}{a subject crosses or lingers at a visible safety or security boundary without authorization} &
\suppbanklist{an authorized worker guard resident or responder occupies the area for a routine purpose}{a pedestrian or vehicle follows an allowed crossing turn lane or waiting area}{a visitor courier tourist or inspector follows a visible routine access path} \\
Environmental, structural, and industrial hazards &
\suppbanklist{a structure object electrical source or machine visibly fails and threatens a person}{flood severe weather debris or structural damage creates immediate visible danger}{machinery equipment collapse or a falling object visibly traps strikes or crushes a person} &
\suppbanklist{workers operate build repair or maintain equipment under controlled conditions}{rain wind reflections flags water or clutter moves without failure or danger}{tools materials machines vehicles and structures remain intact with no visible accident} \\
\bottomrule
\end{tabular}
\caption*{\textbf{Paraphrase B bank.} Complete frozen hazard, mechanism-specific benign, and generic-normal descriptions used in the textual-realization control.}
\end{table*}

\let\suppbanklist\relax

\clearpage